\documentclass[runningheads]{llncs}
\usepackage{graphicx}
\usepackage{comment}
\usepackage{amsmath,amssymb} 
\usepackage{color}
\usepackage{caption}
\usepackage{breakcites}
\usepackage{subcaption}

\usepackage{algorithm}
\usepackage{algpseudocode}
\usepackage[normalem]{ulem}
\usepackage{makecell}
\usepackage{rotating}
\usepackage{booktabs}
\usepackage{gensymb}
\usepackage{wrapfig}
\usepackage{enumitem}
\usepackage{float}
\usepackage[dvipsnames,svgnames,x11names]{xcolor}
\newcommand{\RNum}[1]{\uppercase\expandafter{\romannumeral #1\relax}}

\usepackage[pagebackref=true,breaklinks=true,letterpaper=true,colorlinks,bookmarks=false,citecolor=blue,linkcolor=blue]{hyperref}

\newcommand{\norm}[1]{\left\lVert#1\right\rVert}

\long\def\IGNORE#1{} \long\def\COMMENT#1{}

\begin{document}
\pagestyle{headings}
\mainmatter
\def\ECCVSubNumber{2253}  

\def\authornote#1#2#3{{\textcolor{#2}{\textsl{\small[#1: #3]}}}}

\newcommand{\xlnote}[1]{\authornote{XT}{magenta}{#1}} 
\newcommand{\slnote}[1]{\authornote{SL}{Red}{#1}} 
\newcommand{\khnote}[1]{\authornote{KH}{Blue}{#1}} 
\newcommand{\sdnote}[1]{\authornote{SD}{Green}{#1}} 
\newcommand{\JK}[1]{\authornote{JK}{DarkOrange}{#1}} 

\title{Self-supervised Single-view 3D Reconstruction via Semantic Consistency} 

\titlerunning{Self-supervised Single-view 3D Reconstruction via Semantic Consistency}
%

\author{Xueting Li\inst{1} \and
Sifei Liu\inst{2} \and
Kihwan Kim\inst{2} \and
Shalini De Mello\inst{2} \and
Varun Jampani\inst{2} \and
Ming-Hsuan Yang\inst{1} \and
Jan Kautz\inst{2}}
\authorrunning{Li et al.}
%
\institute{University of California, Merced \and
NVIDIA }
\maketitle
\begin{center}
    \centering
    \includegraphics[width=0.9\textwidth]{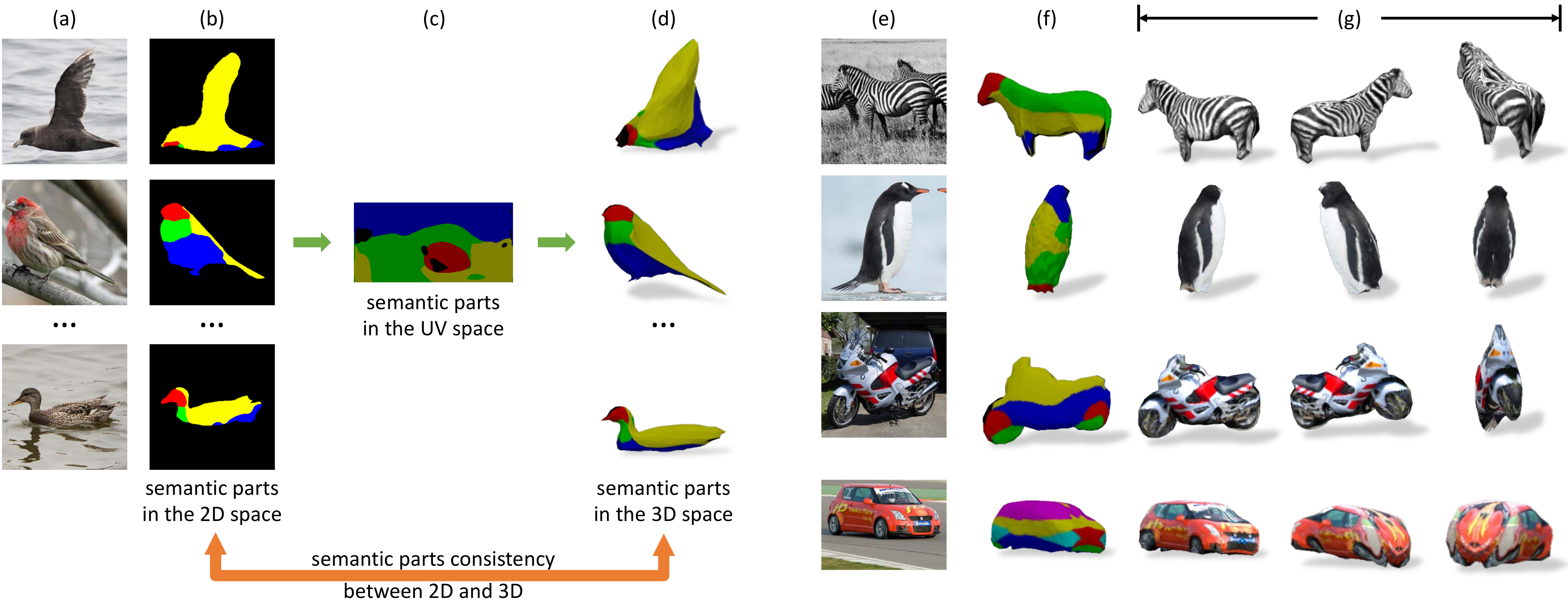}
    \captionof{figure}{\small{\textbf{Self-supervision with semantic part consistency (a--d):} (a) Images of different objects in the same category (e.g., birds in this example) (b) Semantic part segmentation for each image learned via self-supervision. (c) Canonical semantic UV map for the category. (d) Semantic part segmentation on meshes. \textbf{Single-view 3D Mesh reconstruction (e--g):} Reconstruction (inference) of each single-view image (e) is demonstrated in (g), along with semantic labels of the mesh in (f).}}
    \label{fig:teaser}
\end{center}

\vspace{-6mm}
\begin{abstract}
We learn a self-supervised, single-view 3D reconstruction model that predicts the 3D mesh shape, texture and camera pose of a target object with a collection of 2D images and silhouettes. The proposed method does not necessitate 3D supervision, manually annotated keypoints, multi-view images of an object or a prior 3D template. The key insight of our work is that objects can be represented as a collection of deformable parts, and each part is semantically coherent across different instances of the same category (e.g., wings on birds and wheels on cars). Therefore, by leveraging self-supervisedly learned part segmentation of a large collection of category-specific images, we can effectively enforce semantic consistency between the reconstructed meshes and the original images. This significantly reduces ambiguities during joint prediction of shape and camera pose of an object, along with texture. To the best of our knowledge, we are the first to try and solve the single-view reconstruction problem without a category-specific template mesh or semantic keypoints. Thus our model can easily generalize to various object categories without such labels, e.g., horses, penguins, etc. Through a variety of experiments on several categories of deformable and rigid objects, we demonstrate that our unsupervised method performs comparably if not better than existing category-specific reconstruction methods learned with supervision. Codes and other resources will be released at \url{https://sites.google.com/view/unsup-mesh/home}.
\vspace{-2mm}
\end{abstract}

\vspace{-4mm}
\section{Introduction}
\vspace{-2mm}
\label{sec:introduction}
Recovering both 3D shape and texture, and camera pose from 2D images is a highly ill-posed problem due to its inherent ambiguity.
Existing methods resolve this task by utilizing various forms of supervision such as ground truth 3D shapes~\cite{choy20163d,wen2019pixel2mesh++,wang2018pixel2mesh}, 2D semantic keypoints~\cite{cmrKanazawa18}, shading~\cite{henderson2018learning}, category-level 3D templates~\cite{kulkarni2019csm} or multiple views of each object instance~\cite{NIPS2016_6206,kato2018renderer,wiles2017silnet,rezende2016unsupervised}.
%
These types of supervision signals require tedious human effort, and hence make it challenging to generalize to many object categories that lack such annotations.
%
On the other hand, learning to reconstruct by not using any 3D shapes, templates, or keypoint annotations, i.e., with only a collection of single-view images and silhouettes of object instances, remains challenging. This is because the reconstruction model learned without the aforementioned supervisory signals leads to erroneous 3D reconstructions. A typical failure case is caused by the ``camera-shape ambiguity'', wherein, incorrectly predicted camera pose and shape result in a rendering and object boundary that closely match the input 2D image and its silhouette, as shown in Figure~\ref{fig:com} (c) and (d).
%

Interestingly, humans, even infants who have never been taught about objects in a category, tend to mentally reconstruct objects in that category by perceiving them as a combination of several basic parts, e.g., a bird has two legs, two wings, and one head, etc., and use the parts to associate all the divergent instances of the category.
%
By observing object parts, humans can also roughly infer the relative camera pose and 3D shape of any specific instance.
In computer vision, a similar intuition is formulated by the deformable parts model, where objects are represented as a set of parts arranged in a deformable configuration~\cite{felzenszwalb2009object,pepik20123d}. 

Inspired by this intuition, we learn a single-view reconstruction model from a collection of images and silhouettes. We utilize the semantic parts in both the 2D and 3D space, along with their consistency to correctly estimate shape and camera pose.
Specifically, we first leverage self-supervised co-part segmentation (SCOPS~\cite{hung:CVPR:2019}) to decompose 2D images into a collection of semantic parts (Figure~\ref{fig:teaser}(b)). 
By exploiting the property of \textit{semantic part invariance}, which states that the semantic part label of a point on the mesh surface does not change even when the mesh shape is deformed, we associate the semantic parts of \textit{different} object instances with each other and build a category-level canonical semantic UV map (Figure~\ref{fig:teaser}(c)). 
The semantic part label of each point on the reconstructed mesh surface (Figure~\ref{fig:teaser}(d)) is then defined by this canonical semantic UV map.
%
Finally, we resolve the aforementioned ``camera-shape ambiguity'' and learn the self-supervised reconstruction model by encouraging the consistency of semantic part labels in both the 2D and 3D space (Figure~\ref{fig:teaser}, orange arrow). Furthermore, we train our model by iteratively learning (a) instance-level reconstruction and (b) a category-level template mesh from scratch. Thus, our model also does not require a pre-defined 3D template mesh or any other shape prior.
Our main contribution is a 3D reconstruction model that is able to:
\begin{itemize}
    \vspace{-2mm}
	\item Conduct single-view mesh reconstruction \textit{without} any of the following forms of supervision: category-level 3D template prior, annotated keypoints, camera pose or multi-view images. In other words, the model can be generalized to other categories which do not have well-defined keypoints, e.g., penguin.
	\item Leverage the \textit{semantic part invariance} property of object instances of a category as a deformable parts model.
	\item Learn a category-level 3D shape template from scratch via iterative learning.
	\item Perform comparably to the state-of-the-art supervised methods~\cite{cmrKanazawa18,kulkarni2019csm} trained with either pre-defined templates or annotated keypoints, while also improving the self-supervised semantic co-part segmentation model (SCOPS~\cite{hung:CVPR:2019}).
\end{itemize}

\vspace{-4mm}
\section{Related Work}
\vspace{-2mm}
\subsubsection{3D Shape Representation}
Various representations have been explored for 3D processing tasks, including point clouds~\cite{fan2017point}, implicit surfaces~\cite{Occupancy_Networks,liu2019_implit}, triangular meshes~\cite{cmrKanazawa18,kato2018renderer,liu2019softras,kato2019vpl,wang2018pixel2mesh,pan2019deep,wen2019pixel2mesh++} and voxel grids~\cite{choy20163d,Girdhar16b,gwak2017weakly,drcTulsiani17,wiles2017silnet,NIPS2016_6206,zhu2017rethinking,hspHane17}.
Among these, while both voxels and point clouds are more friendly to deep learning architectures (e.g., VON~\cite{3dgan,VON}, PointNet~\cite{qi2016pointnet,qi2017pointnetplusplus}, etc), they suffer either from issues of memory inefficiency or are not amenable to differentiable rendering.
Hence, in this work, we adopt triangular meshes~\cite{cmrKanazawa18,kato2018renderer,liu2019softras,kato2019vpl,wang2018pixel2mesh,pan2019deep,wen2019pixel2mesh++} for 3D reconstruction.
\vspace{-4mm}
\subsubsection{Single-view 3D Reconstruction}
Single-view 3D reconstruction~\cite{choy20163d,Girdhar16b,gwak2017weakly,drcTulsiani17,wiles2017silnet,NIPS2016_6206,zhu2017rethinking,fan2017point,henderson2018learning} aims to reconstruct a 3D shape given a single input image. 
One line of works have explored this ill-posed task with varying degree of supervision.
Several methods~\cite{wang2018pixel2mesh,pan2019deep,wen2019pixel2mesh++} utilize image and ground truth 3D mesh pairs as supervision. This either requires significant manual annotation effort~\cite{xiang2016objectnet3d} or is restricted to synthetic data~\cite{shapenet2015}. 
More recently, a few works~\cite{kato2018renderer,liu2019softras,kato2019vpl,chen2019dibrender} avoid 3D supervision by taking advantage of differentiable renderers~\cite{kato2018renderer,liu2019softras,chen2019dibrender} and the ``analysis-by-synthesis" approach, with either multiple views, or known ground truth camera poses. 

To further relax constraints on supervision, Kanazawa et al.~\cite{cmrKanazawa18} explored 3D reconstruction from a collection of images of different instances. However, their method still requires annotated 2D keypoints to infer camera pose correctly. It is also the first work to propose a learnable category-level 3D template shape, which, however, needs to be initialized from a keypoint-dependent 3D convex hull.
Similar problem settings have also been explored in other methods~\cite{szabo2018unsupervised,Wu_2019,henderson2019learning}, but with object categories restricted to rigid or structured objects, such as cars or faces.
Different from all these works, we target both rigid and non-rigid objects (e.g., birds, horses, penguins, motorbikes and cars shown in Figure~\ref{fig:teaser} (e)-(g)) and propose a method that jointly estimates a 3D mesh, texture, and camera pose from a single-view image, using only a collection of images with silhouettes as supervisions. In other words, we do not require 3D template priors, annotated keypoints, or multi-view images.
\vspace{-4mm}
\subsubsection{Self-supervised Correspondence Learning}
Our work is also related to self-supervised cross-instance correspondence learning, via landmarks~\cite{thewlis2017unsupervised,zhang2018unsupervised,honari2018improving,simon2017hand}, part segments~\cite{collins2018,hung:CVPR:2019}, or canonical surface mapping~\cite{kulkarni2019csm}.
We utilize self-supervised co-parts segmentation~\cite{hung:CVPR:2019} to enforce semantic consistency, which was originally proposed purely for 2D images.
%
The work of~\cite{kulkarni2019csm} learns a mapping function that maps pixels in 2D images to a predefined category-level template in a self-supervised manner. However, it dose not use the learned correspondence for 3D reconstruction.
We show that our work, despite having a focus on 3D reconstruction, outperforms~\cite{kulkarni2019csm} at learning 2D to 3D correspondences as well.

\vspace{-2mm}
\section{Approach}
\vspace{-2mm}
\label{sec:approach}
\begin{figure}[t]
\centering
\includegraphics[width=.9\linewidth]{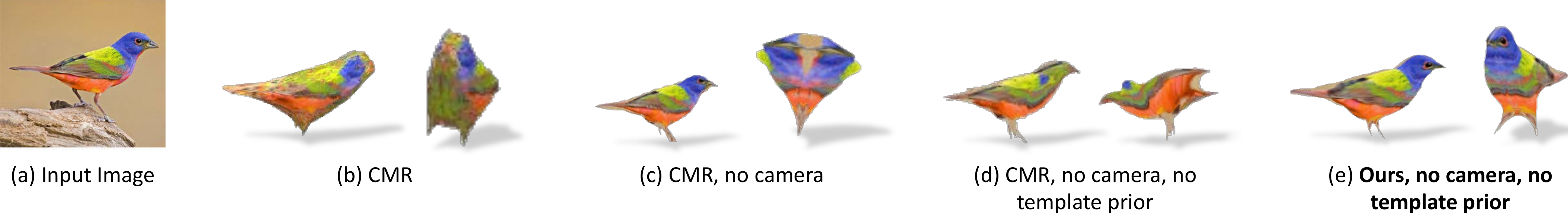}
\vspace{-2mm}
\caption{\small{\textbf{Comparison with baselines.} 
Each reconstructed mesh is rendered in the original view of the input image and the frontal view of the bird. 
(b) shows the result from CMR with camera pose and template prior supervision, (c) shows CMR with only template prior, and (d) shows CMR without both supervisions where the model completely fails to learn the texture and shape. In contrast, our model in (e) reconstructs correctly even without the supervision from both camera pose and template prior.}
\IGNORE{
When camera supervision is removed as shown in (c), the model reconstructs a mesh that looks plausible only from the predicted camera pose. If the template is further removed as shown in (d), the model completely fails to learn meaningful texture and shape. In contrast, our model reconstructs correctly even without camera and template supervision (e).
} 
}
\vspace{-2mm}
\label{fig:com}
\end{figure}

To fully reconstruct the 3D mesh of an object instance from an image, a network should be able to jointly predict the shape and texture of the object, and the camera pose of the image. 
We start with the existing network from~\cite{cmrKanazawa18} (CMR) as the baseline reconstruction network.
Given an input image, CMR extracts the image features using an encoder $E$ and jointly predicts the mesh shape, camera pose and mesh texture by three decoders $D_\mathrm{shape}$, $D_\mathrm{camera}$ and $D_\mathrm{texture}$ . The mesh shape $V$ is reconstructed by predicting vertex offsets $\Delta{V}$ to a category-specific shape template $\bar{V}$, while the camera pose $\theta$ is represented by a weak perspective transformation. 
To reconstruct mesh textures, the texture decoder outputs a UV texture flow ($I_\mathrm{flow}$) that maps pixels from the input image to the UV space. A pre-defined mapping function $\Phi$ further maps each pixel in the UV space to a point on the mesh surface.

One of the key elements for the CMR method to perform well is to exploit \emph{mannually annotated semantic keypoints} for
(i) precisely pre-computing the ground truth camera pose for each instance, and (ii) estimating a category-level 3D template prior.
However, annotating keypoints is tedious, not well-defined for most object categories in the world and impossible to generalize to new categories. 
Thus, we propose a method within a more scalable, but challenging self-supervised setting \emph{without} using manually annotated keypoints to estimate camera pose or a template prior.

Not surprisingly, simply taking out the keypoints supervision, as well as all the related information (i.e., the camera pose and the template prior) from the CMR network makes it unable to predict camera pose and shape correctly, as shown in Figure~\ref{fig:com}(c) and~(d).
This is due to the inherent ambiguity of hallucinating 3D meshes from only single-view 2D observations, where the model trivially picks a combination of camera pose and shape that yields the rendering that matches the given image and silhouette. 
Consider an extreme case, where the model predicts the front view for all instances, but is still able to match the image and silhouette observations by deforming each instance mesh accordingly.

\begin{figure}[t]
    \centering
    \includegraphics[width=0.9\linewidth]{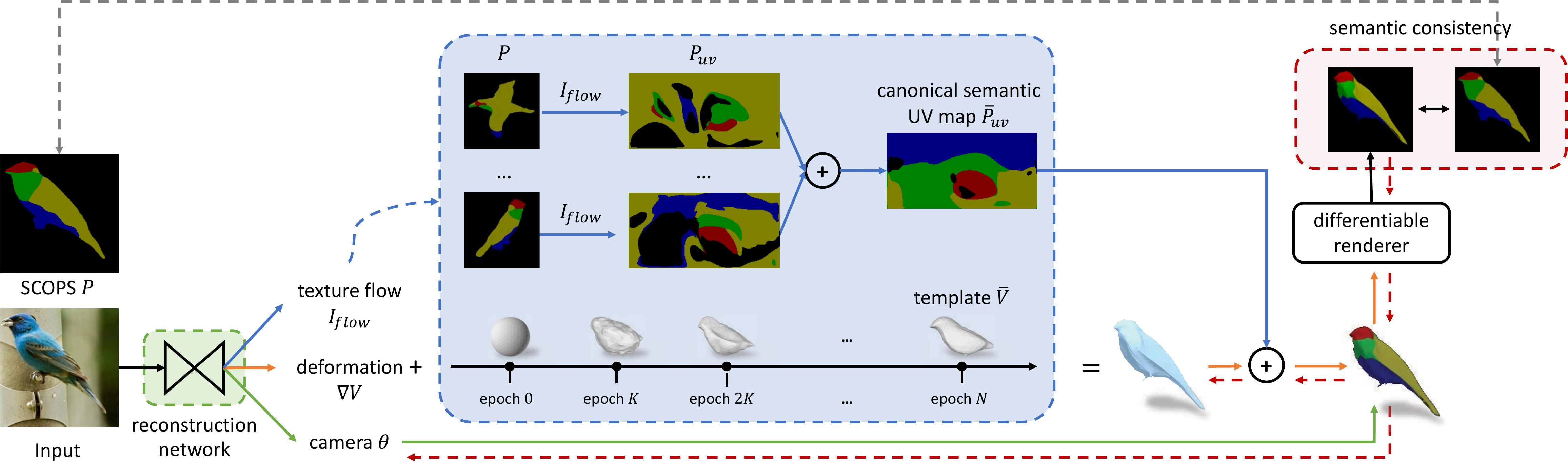}
    {\caption{\small{\textbf{Overview.}(a) Green box: The reconstruction network. (b) Red box: Semantic part consistency constraint, see Section~\ref{sec:semantics} for more details. (c) Blue box: Computing the canonical semantic UV map and the template shape using the reconstruction network, see Section~\ref{sec:progressive}. The red dashed arrows show that the gradients from the semantic part consistency constraint facilitate the camera and shape estimation.
    }\label{fig:overview}}}
    \vspace{-2mm}
\end{figure}
In this work, we propose a framework (Figure~\ref{fig:overview}) designed for self-supervised mesh reconstruction learning, i.e. with only a collection of images and silhouettes as supervision. The framework consists of: (i) A reconstruction network (green box) that has the same architecture as~\cite{cmrKanazawa18} -- it consists of an image encoder $E$ and three decoders $D_\mathrm{shape}$, $D_\mathrm{camera}$ and $D_\mathrm{texture}$ that jointly predict the mesh deformation $\Delta V$, texture flow $I_\mathrm{flow}$ and camera pose $\theta$ for the instance in the image. (ii) A semantic consistency constraint (red box in Figure~\ref{fig:overview}) that regularizes the learning of module (i) and largely resolves the aforementioned ``camera-shape ambiguity'' under the self-supervised setting. We introduce this module in Section~\ref{sec:semantics}. (iii) A module that learns the canonical semantic UV map and category-level template from scratch (blue box in Figure~\ref{fig:overview}). This module is iteratively trained with module (i) and discussed in Section~\ref{sec:progressive}.

\vspace{-2mm}
\subsection{Resolving Camera-Shape Ambiguity via Semantic Consistency}
\vspace{-2mm}
\label{sec:semantics}
In this section, we show the key to solving the ``camera-shape ambiguity'' is to make use of the semantic parts of object instances in both 3D and 2D.
Specifically, we exploit the fact that (i) in the 2D space, the self-supervised co-part segmentation~\cite{hung:CVPR:2019}
provides correct part segments for the majority of the object instances, even for those with large shape variations (see Figure~\ref{fig:teaser}(b)); and (ii) in the 3D space, semantic parts are invariant to mesh deformations, i.e., the semantic part label of a specific point on the mesh surface is consistent across all reconstructed instances of a category. We demonstrate that this \textit{semantic part invariance} allows us to build a category-level semantic UV map, namely the canonical semantic UV map, shared by all instances, which in turn allows us to assign semantic part labels to each point on the mesh. 
By enforcing consistency between the canonical semantic map and an instance's part segmentation in the 2D space, the camera-shape confusion can be largely resolved.

\vspace{-2mm}
\subsubsection{Part Segmentation in 2D via SCOPS~\cite{hung:CVPR:2019}} 
SCOPS is a self-supervised method that learns semantic part segmentation from a collection of images of an object category (see Figure~\ref{fig:teaser}(b)).
The model leverages concentration and equivalence loss functions, as well as part basis discovery to output a probabilistic map w.r.t.\ the discovered parts, which are semantically consistent across different object instances.
In Figure~\ref{fig:exp_scops} (second row), we demonstrate several examples of semantic part segments predicted by the SCOPS. Although SCOPS successfully discovers all semantic parts for most instances, the shape and size of each part is not consistent across different instances, e.g., the part corresponding to the head of the bird in Figure~\ref{fig:exp_scops} (second row) is too small. 
We discuss that, besides generalizing SCOPS for reconstructing objects, how our model is able to improve SCOPS in return, in Section~\ref{sec:scops}.

\vspace{-2mm}
\subsubsection{Part Segmentation in 3D via Canonical Semantic UV Map}
\label{sec:cano_uv}
Given the semantic part segmentation of 2D images estimated by SCOPS, how can we obtain the semantic part labels for each point on the mesh surface? 
One intuitive way is to obtain a mapping from the 2D image space to the 3D shape space.
Therefore, we propose to first utilize the learned texture flow $I_\mathrm{flow}$ by our reconstruction network that naturally forms a mapping from the 2D image space to the UV texture space, and then further map the semantic labels from the UV space to the mesh surface by the pre-defined mapping function $\Phi$. 
We denote the semantic part segmentation of image $i$ as $P^i\in\mathbb{R}^{H\times W\times N_p}$ (see Figure~\ref{fig:overview} in the blue bbox), where $H$, $W$ are the height and width of the image and $N_p$ is the number of semantic parts.
By mapping $P^i$ from the 2D image space to the UV space using the learned texture flow, we obtain a ``semantic UV map'' denoted as $P_\mathrm{uv}^i\in\mathbb{R}^{H_\mathrm{uv}\times W_\mathrm{uv}\times N_p}$, where $H_\mathrm{uv}$ and $W_\mathrm{uv}$ are the UV map's height and width, respectively.

Ideally, all instances should result in the same semantic UV map -- the canonical semantic UV map for a category, regardless of shape differences of instances.
This is because: (i) the \textit{semantic part invariance} states that the semantic part labels assigned to each point on the mesh surface are consistent across different instances; and 
(ii) the mapping function $\Phi$ that maps pixels from the UV space to the mesh surface is pre-defined and independent of deformations in the 3D space, such as face location or area changes.
Thus, the semantic part labels of pixels in the UV map should also be consistent across different instances.

However, if we directly sample the individual $P^i$ via the learned texture flow $I_\mathrm{flow}$, the obtained semantic UV maps are indeed very different between instances, as shown in Figure~\ref{fig:overview} (blue box). This is caused by the fact that (i) the part segmentation predictions produced by the self-supervised SCOPS method are noisy, and (ii) texture flow prediction is highly uncertain for the invisible faces of the reconstructed mesh.
Therefore, we approximate the canonical semantic UV map, denoted as $\bar{P}_\mathrm{uv}$ by aggregating the individual semantic UV maps:
\vspace{-2mm}
\begin{equation}\vspace{-2mm}
\small{
\label{eq:template_semantic}
    \bar{P}_\mathrm{uv} = \frac{1}{|\mathcal{U}|}\sum_{i\in \mathcal{U}} I_\mathrm{flow}^i(P^i),}
\end{equation}
where $I_\mathrm{flow}^i(P^i)$ indicates the sampling of $P^i$ by $I_\mathrm{flow}$ and $\mathcal{U}$ is a subset of selected training samples with accurate texture flow prediction (the selection process can be found in the appendix). Through this aggregation process, $\bar{P}_\mathrm{uv}$ produces a mean semantic UV map, which effectively eliminates outliers (i.e., instances with incorrect SCOPS), as well as the noisy pixel-level predictions.

\vspace{-4mm}
\subsubsection{Semantic Consistency between 2D and 3D}

\begin{figure}
\centering
\begin{minipage}{.48\textwidth}
  \centering
  \includegraphics[height=0.45\textwidth]{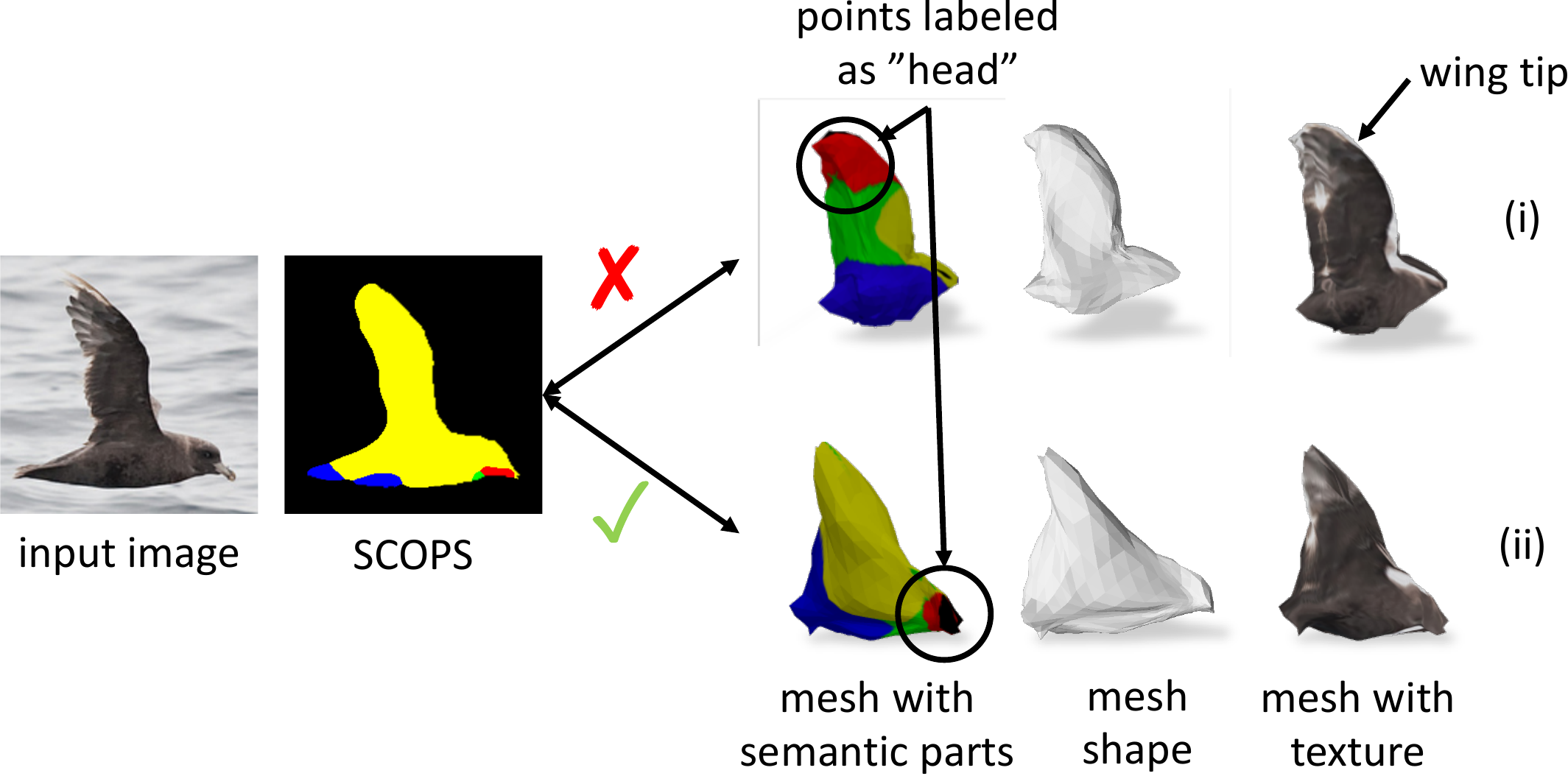}
  \captionof{figure}{\small{\textbf{Semantic part invariance.} (i) Incorrect reconstruction without semantic part consistency. (ii) Reconstruction with consistency.}}
  \label{fig:semantic_constraint}
\end{minipage}%
\hfill
\begin{minipage}{.48\textwidth}
  \centering
  \includegraphics[height=0.45\textwidth]{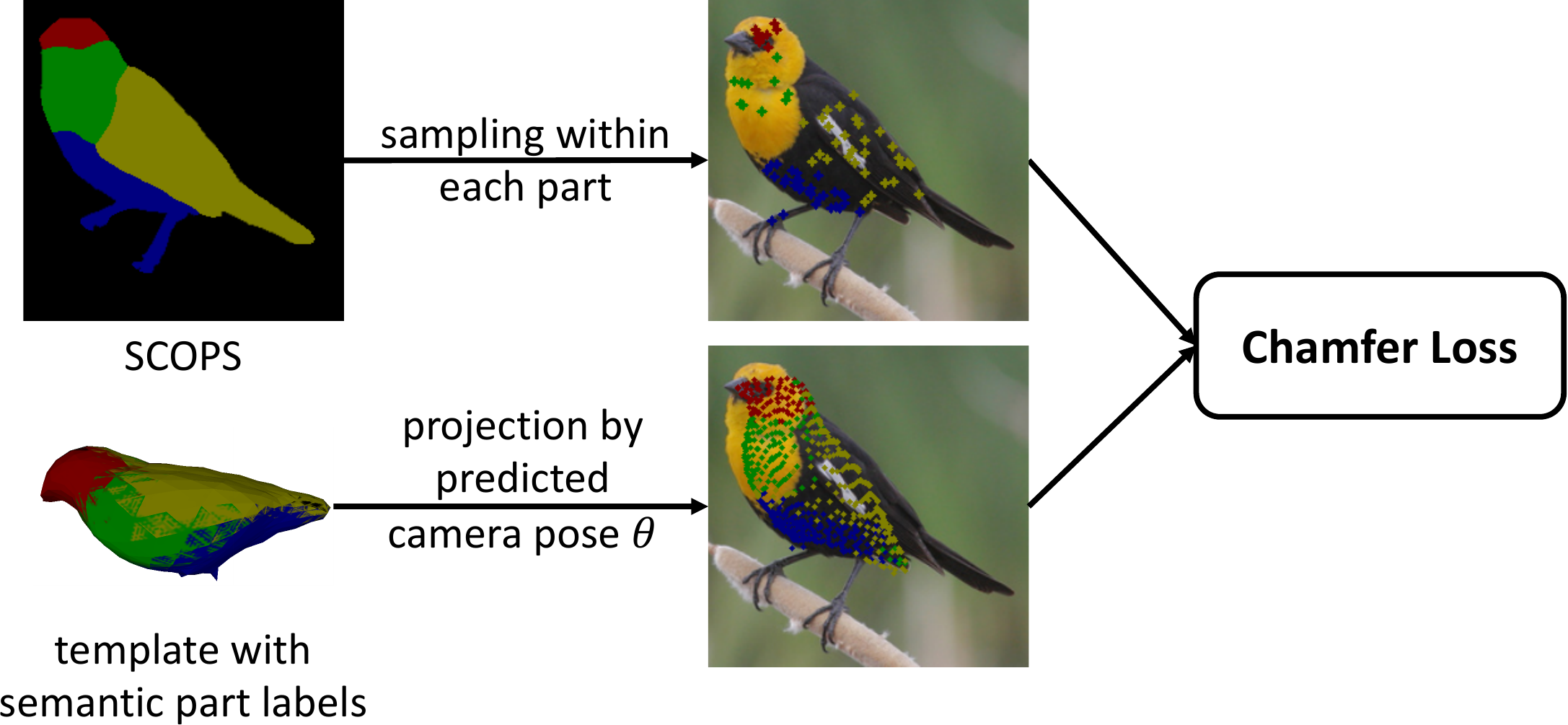}
  \captionof{figure}{\small{\textbf{Visualization of the vertex-based semantic consistency constraint.} Points of the same color belong to the same semantic part.}}
  \label{fig:vertex}
\end{minipage}
\vspace{-2mm}
\end{figure}

As mentioned above, because the self-supervisedly learned model only relies on images and silhouettes, which do not provide any semantic part information, the model suffers from the ``camera-shape ambiguity'' introduced in Section~\ref{sec:introduction}.
Take row (i) in Figure~\ref{fig:semantic_constraint} as an example. The model erroneously forms the wing tip in the reconstructed bird by deforming faces assigned as the ``head part'' (colored in red). This incorrect shape reconstruction, associated with an incorrect camera pose, however, can yield a rendering that matches the image and silhouette observation.

This ambiguity, although is not easy to spot by only comparing the rendered reconstruction image with the input image, however, can be identified once the semantic part label for each point on the mesh surface is available.
One can tell that the reconstruction in row (i) of Figure~\ref{fig:semantic_constraint} is wrong by comparing the rendering of the semantic part labels on the mesh surface and the 2D SCOPS part segmentation. 
Only when the camera pose and shape are both correct, will the rendering and the SCOPS segmentation be consistent, as shown in row (ii) in Figure~\ref{fig:semantic_constraint}.
This observation inspires us to propose a probability and a vertex-based constraint that facilitate camera pose and shape learning by encouraging the consistency of semantic part labels in both 2D images and the mesh surface.

\vspace{-2mm}
\paragraph{\textbf{Probability-based constraint.}}
For each reconstructed mesh instance $i$, we map the canonical semantic UV map $\bar{P}_\mathrm{uv}$ onto its surface by the UV mapping $\Phi$ and render it using the predicted camera pose $\theta^i$. We denote the projection from 3D to 2D as $\mathcal{R}$. 
We constrain the projected probability map to be close to the SCOPS part segmentation probability map $P^i$ by computing the loss:
\vspace{-2mm}
\begin{equation}\vspace{-2mm}
    \label{eq:sem_face}
    L_\mathrm{sp} = \norm{P^i - \mathcal{R}(\Phi(\bar{P}_\mathrm{uv}); \theta^i)}^2.
\end{equation}
We empirically found the mean squared error (MSE) metric to be more robust than the Kullback–Leibler divergence for comparing two probability maps.
%
\vspace{-2mm}
\paragraph{\textbf{Vertex-based constraint.}} We also propose a vertex-based constraint to enhance semantic part consistency (Figure~\ref{fig:vertex}) by enforcing that 3D vertices assigned a part label $p$, after being projected to the 2D domain with the predicted camera pose $\theta^i$, align with the area assigned to that part in the input image:
\vspace{-2mm}
\begin{equation}\vspace{-2mm}
    \label{eq:sem_vertex}
    \small{
    L_\mathrm{sv} = \sum_{p=1}^{N_p} \frac{1}{|\bar{V}_p|}\mathrm{Chamfer}(\mathcal{R}(\bar{V}_p;\theta^i), Y_p^i)},
\end{equation}
where $\bar{V}_p$ is the set of vertices on a learned category-level 3D \textit{template} $\bar{V}$ (see Section~\ref{sec:progressive}) with the part label $p$, $Y_p^i$ is the set of 2D pixels sampled from the part $p$ in the original input image and $N_p$ is the number of parts.
Here we use the \textit{Chamfer distance}, because the projected vertices and pixels with the same part label $p$ in the input image do not have a strictly one-to-one correspondence.

Note that, $\bar{V}_p$ is a set of vertices on the category-level shape template $\bar{V}$ as opposed to each instance reconstruction $V^i$, since using ${V^i}$ results in a degenerate solution where the network only alters 3D shape to satisfy this vertex-based constraint, rather the camera pose. Instead, using $\bar{V}$ drives the network towards learning the correct camera pose, in addition to shape.

\vspace{-2mm}
\subsection{Progressive Training}
\vspace{-2mm}
\label{sec:progressive}

We train the framework in Figure~\ref{fig:overview} by a progressive training method based on two considerations:
(a) building the canonical semantic UV map, introduced in Section~\ref{sec:semantics}, requires reliable texture flows to map the SCOPS from 2D images to the UV space. Thus the canonical semantic UV map can only be obtained after the reconstruction network is able to predict texture flow reasonably well, and 
(b) a canonical 3D shape template~\cite{cmrKanazawa18,kulkarni2019csm} is desirable, since it speeds up the convergence of the network~\cite{cmrKanazawa18} and also avoids degenerate solutions when applying the \textit{vertex-based constrain} as introduced in Section~\ref{sec:semantics}. However, jointly learning the category-level 3D shape template and the instance-level reconstruction network leads to undesired trivial solutions. Thus, we propose an expectation-maximization (EM) style progressive training procedure below. In the E-step, we train the reconstruction network with the current template and canonical semantic UV map fixed, and in the M-step, we update the template and the canonical semantic UV map using the reconstruction network learned in the E-step.

\vspace{-2mm}
\subsubsection{E-step: Learning Instance-specific Reconstruction}
In the E-step, we fix the canonical semantic UV map as well as the category-level template and train the reconstruction network mainly with the following objectives. (i) A negative IoU objective~\cite{kato2019vpl} between the rendered and the ground truth silhouettes for shape learning. (ii) A perceptual distance objective~\cite{zhang2018perceptual,cmrKanazawa18} between the rendered and the input RGB images for texture learning. (iii) The probability and vertex-based constraints introduced in Section~\ref{sec:semantics} to resolve the ``camera-shape ambiguity'' under the self-supervised setting. (iv) A texture consistency constraint to facilitate accurate texture flow learning that will be introduced in Section~\ref{sec:tcyc}. Other constraints is included in the appendix. Note that in the first E-step, the template is a sphere and the probability as well as vertex-based constraints are not involved.

\vspace{-2mm}
\subsubsection{M-step: Canonical UV Map and Template Learning}
In the M-step, we compute the canonical semantic UV map as introduced in Section~\ref{sec:semantics} and learn a category-level template from scratch, i.e., from a sphere primitive.
As far as we know, we are the first method that learns a category-level template from scratch.
This is in contrast to existing methods~\cite{kulkarni2019csm}, where the template is either a readily available instance mesh from the category or estimated from annotated keypoints~\cite{cmrKanazawa18}.
Jointly learning the shape template along with the reconstruction network does not guarantee a meaningful ``mean shape'' which encapsulates the most representative characteristics of objects in a category. Instead, we propose a feed-forward template learning approach: 
the template starts out as a sphere and is updated every $K$ training epochs by:
\vspace{-2mm}
\begin{equation}\vspace{-2mm}
\small{
\label{eq:template_shape}
    \bar{V}_{t} = \bar{V}_{t-1} + D_\mathrm{shape}( \frac{1}{|\mathcal{Q}|}\sum_{i\in \mathcal{Q}} E(I^i))},
\end{equation}
where $\bar{V}_{t}$ and $\bar{V}_{t-1}$ are the updated and current templates, respectively,  $I^i$ is the input image passed to the image encoder $E$ and $D_\mathrm{shape}$ is the shape decoder (see the beginning of Section~\ref{sec:approach}). $\mathcal{Q}$ is a set of selected samples with consistent mesh predictions and the selection process is discussed in the appendix. 
The template $\bar{V}_t$ is the mean shape of instances in a category for the current epoch, which enforces a meaningful shape (e.g., the template looks like a bird) rather than an arbitrary form for the category.

\vspace{-2mm}
\subsection{Texture Cycle Consistency Constraint}\vspace{-2mm}
\label{sec:tcyc}

\begin{figure}[t]
\centering
\begin{subfigure}{.45\textwidth}
  \centering
  \includegraphics[height=0.5\textwidth]{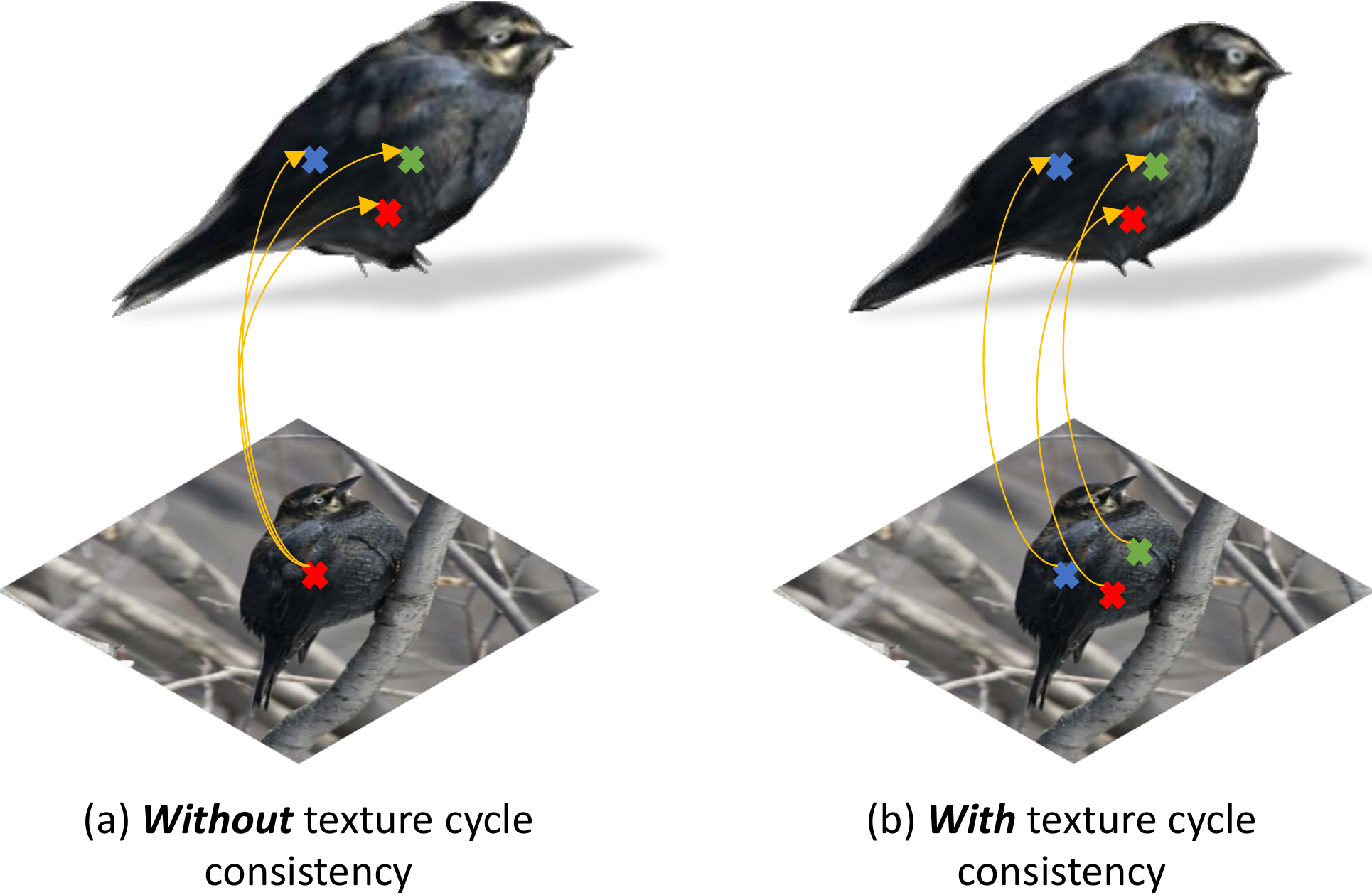}
\end{subfigure}
\begin{subfigure}{.45\textwidth}
  \centering
  \includegraphics[height=0.5\textwidth]{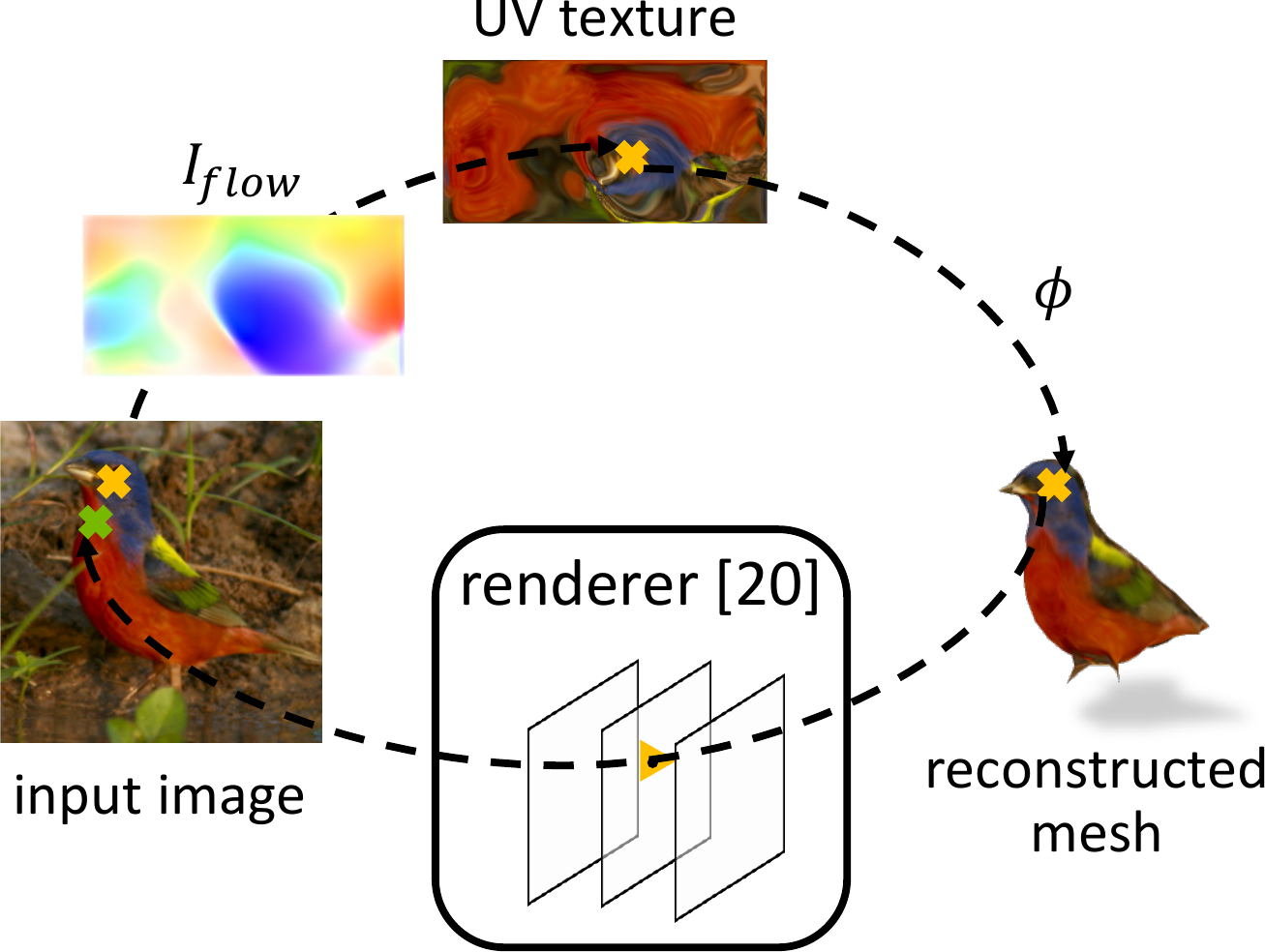}
\end{subfigure}%
\caption{\textbf{Visualization of the texture cycle consistency constraint.} Left: Visualization of the effectiveness of the texture cycle consistency constraint. Right : The process of  texture cycle consistency constraint computation. }
\vspace{-2mm}
\label{fig:tcyc}
\end{figure}

As shown in Figure~\ref{fig:tcyc} (Left), one issue with the learned texture flow is that the texture of 3D mesh faces with a similar color (e.g., black) can be incorrectly sampled from a single pixel location of the image.
Thus we introduce a texture cycle consistency objective to regularize the predicted texture flow (i.e., 2D$\rightarrow$3D) to be consistent with the camera projection (i.e., 3D$\rightarrow$2D).
As shown in Figure~\ref{fig:tcyc} (Right), considering the pixel marked with a yellow cross in the input image, it can be mapped to the mesh surface through the predicted texture flow $I_{flow}$ along with the pre-defined mapping function $\Phi$ introduced in Section~\ref{sec:approach}. Meanwhile, its mapping on the mesh surface can be re-projected back to the 2D image by the predicted camera pose, as shown by the green cross in Figure~\ref{fig:tcyc} (Right). If the predicted texture flow conforms to the predicted camera pose, the yellow and green crosses would overlap, forming a $2D\rightarrow3D\rightarrow2D$ cycle.

Formally, given a triangle face $j$, we denote the set of input image pixels mapped to this face by texture flow as $\Omega_\mathrm{in}^{j}$.
We further infer the set of pixels (denoted as $\Omega_\mathrm{out}^{j}$) projected from the triangle face $j$ in the rendering operation by taking advantage of the probability map $\mathcal{W}\in\mathcal{R}^{|F|\times (H\times W)}$ in the differentiable renderer~\cite{liu2019softras} 
where $|F|, H, W$ are the number of faces, height and width of the input image, respectively. Each entry in $\mathcal{W}_j^m$ indicates the probability of face $j$ being projected onto the pixel $m$.
We compute the geometric center of both sets ($\Omega_\mathrm{in}^{j}$ and $\Omega_\mathrm{out}^{j}$), denoted by $\mathcal{C}_\mathrm{in}^j$ and $\mathcal{C}_\mathrm{out}^j$, respectively as:
%
\vspace{-2mm}
\begin{equation}
    \label{tcyc-center}
     \vspace{-0.5mm}
     \small {
     \mathcal{C}_\mathrm{in}^j = \frac{1}{N_c}\sum_{m=1}^{N_c}\Phi(I_\mathrm{flow}(\mathcal{G}^m))_j
     ;\quad \mathcal{C}_\mathrm{out}^j = \frac{\sum_{m=1}^{H\times W} \mathcal{W}_j^m\times \mathcal{G}^m}{\sum_{m=1}^{H\times W} \mathcal{W}_j^m}},
 \end{equation}
where $\mathcal{G}\in \mathbb{R}^{(H\times W)\times 2}$ 
is a standard coordinate grid of the projected image (containing pixel location $(u, v)$ values), and $\Phi$ is the  
fixed UV mapping that, along with the texture flow $I_\mathrm{flow}$ maps pixels from the 2D input image to a mesh face $j$, as discussed in the beginning of Section~\ref{sec:approach}. $N_c$ is the number of pixels in the input image mapped to each triangular face and $\times$ indicates multiplication between two scalars.
We constrain the predicted texture flow to be consistent with the rendering operation by encouraging $\mathcal{C}_\mathrm{in}^j$ to be close to $\mathcal{C}_\mathrm{out}^j$:
\vspace{-3mm}
\begin{equation}\vspace{-2mm}
    \small{
    \label{tcyc}
    L_\mathrm{tcyc} = \frac{1}{|F|}\sum_{j=1}^{|F|}\norm{\mathcal{C}_\mathrm{in}^j - \mathcal{C}_\mathrm{out}^j}_F^2.}
\end{equation}
We note that while not targeting 3D mesh reconstruction directly, a similar intuition, but with a different formulation was also introduced in~\cite{kulkarni2019csm}.

\vspace{-2mm}
\subsection{Better Part Segmentation via Reconstruction}\vspace{-2mm}
\label{sec:scops}
The proposed 3D reconstruction model can, in turn, be used to improve learning of self-supervised part segmentation~\cite{hung:CVPR:2019} (see Figure~\ref{fig:exp_scops}).
The key intuition is that the category-level canonical semantic UV map $\bar{P}_\mathrm{uv}$ learned in Section~\ref{sec:cano_uv} largely reduces noise in instance-based semantic UV maps. When combined with instance mesh reconstruction and camera pose, it provides reliable supervision for the SCOPS method.

By mapping the canonical UV map to the surface of each reconstructed mesh and rendering it with the predicted camera pose, we obtain psuedo ``ground truth'' segmentation maps as supervision for SCOPS training.
We use the semantic consistency constraint in Section~\ref{sec:semantics} as a measurement to  select the reliable reconstructions with high semantic consistency (i.e., with low probability and vertex-based semantic consistency loss values) to train SCOPS with.  
The improved SCOPS can, in turn, provide better regularization for our mesh reconstruction network, forming an iterative and collaborative learning loop.

\vspace{-2mm}
\section{Experimental Results}
\vspace{-2mm}
We first introduce our experimental settings in Section~\ref{sec:exp_settings}, and present qualitative evaluations for the bird, horse, motorbike and car categories in Section~\ref{sec:exp_quali}. Quantitative evaluations and ablation studies for the contribution of each proposed module are discussed in Section~\ref{sec:exp_quanti} and Section~\ref{sec:exp_ablation}, respectively.

\vspace{-2mm}
\subsection{Experimental Settings}
\label{sec:exp_settings}
\subsubsection{Datasets}
We validate our method on both rigid objects, i.e., \textit{car} and \textit{motorcycle} images from the PASCAL3D+ dataset~\cite{xiang2014beyond}, and non-rigid objects, i.e., \textit{bird} images from the CUB-200-2011 dataset~\cite{wah2011caltech}, \textit{horse}, \textit{zebra}, \textit{cow} images from the ImageNet dataset~\cite{deng2009imagenet} and \textit{penguin} images from the OpenImages dataset~\cite{kuznetsova2018open}.
%
%
%
%

\vspace{-2mm}
\subsubsection{Network Training}
We use a progressive training approach (Section~\ref{sec:progressive}) to learn the model parameters. 
In each E-step, the reconstruction network is trained for 200 epochs and then used to update the template and the canonical semantic UV map in the M-step. 
The only exception is in the first round (a round consists of one E and M-step), where we train the reconstruction network without the semantic consistency constraint. This is because, at the beginning of training, $I_\mathrm{flow}$ is less reliable, which in turn makes the canonical semantic UV map less accurate.
%
%
\vspace{-2mm}
\subsection{Qualitative Results}\vspace{-2mm}
Thanks to the self-supervised setting, our model is able to learn from a collection of images and silhouettes (e.g., horse and cow images~\cite{deng2009imagenet} and penguin images~\cite{kuznetsova2018open}), which cannot be achieved by existing methods~\cite{cmrKanazawa18,wang2018pixel2mesh,yan2016perspective,kato2018renderer} that require extra supervisory signals. 
\label{sec:exp_quali}
\begin{figure*}[t]
    \centering
    \includegraphics[width=0.95\linewidth]{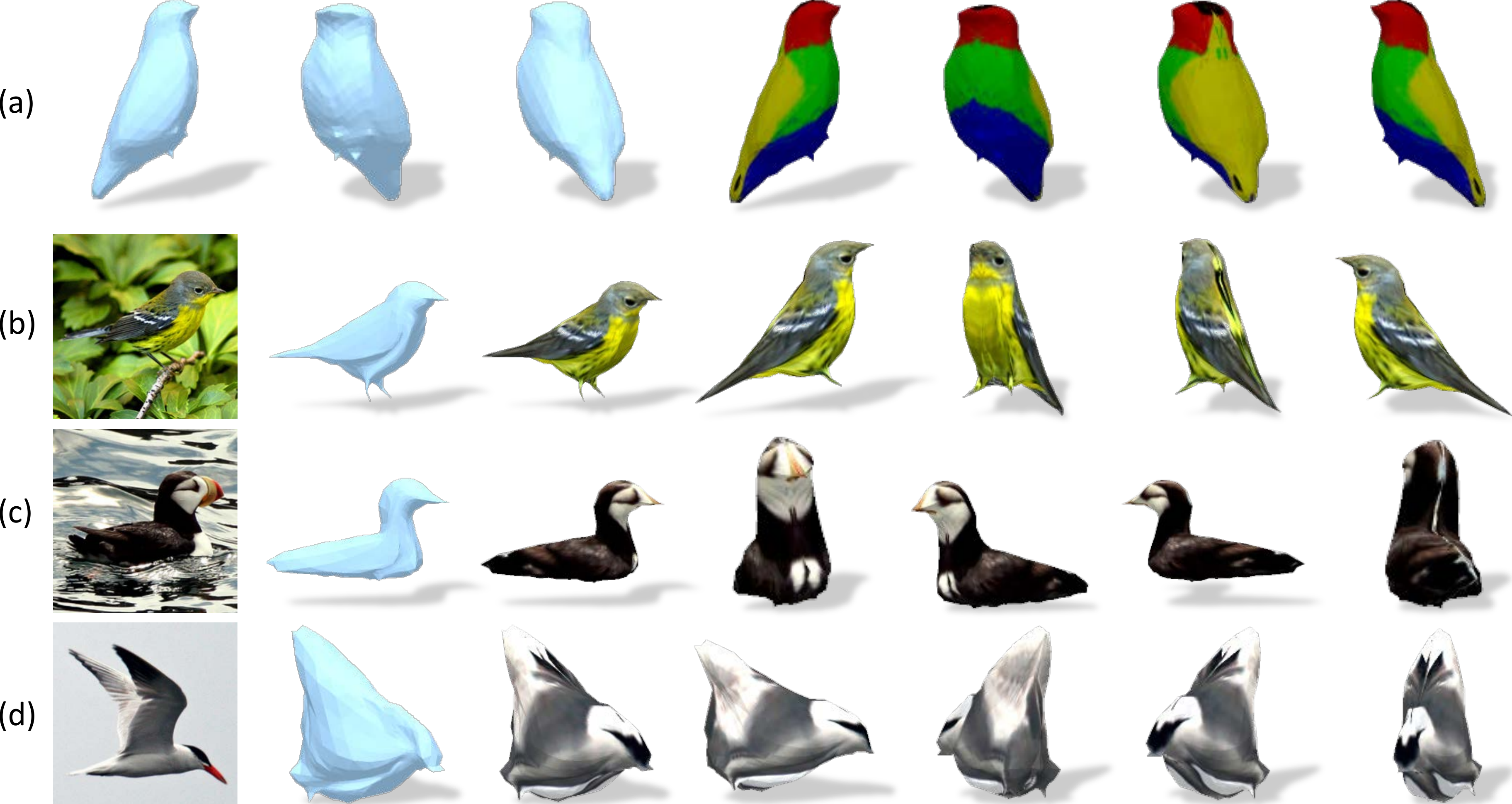}
    {\caption{\small{ \textbf{Learned template and instance reconstructions from single-view images}. (a) The learned template shape (first three columns) and semantic parts (last four columns). (b)-(d) 3D reconstruction from a single-view image. In each row from left to right, we show the input image, reconstruction rendered using the predicted camera view and from four other views. Please see the results for additional views in the appendix video.}}\label{fig:exp_cub}} 
    \vspace{-.2cm}
\end{figure*}

\vspace{-2mm}
\subsubsection{Template and Semantic Parts on 3D Meshes}
We show the learned templates for the bird, horse, motorbike and car categories in Figure~\ref{fig:exp_cub} and Figure~\ref{fig:other_results}, which capture the shape characteristics of each category, including the details such as the beak and feet of a bird, etc.
%
We also visualize the canonical semantic UV map by showing the semantic part labels assigned to each point on the template surface. For instance, the bird meshes have four semantic parts -- head (red), neck (green), belly (blue) and back (yellow) in Figure~\ref{fig:exp_cub}, which are consistent with the part segmentation predicted by the SCOPS method~\cite{hung:CVPR:2019}.

\vspace{-2mm}
\subsubsection{Instance 3D Reconstruction}
\begin{figure}[t]
    \centering
    \includegraphics[width=0.95\linewidth]{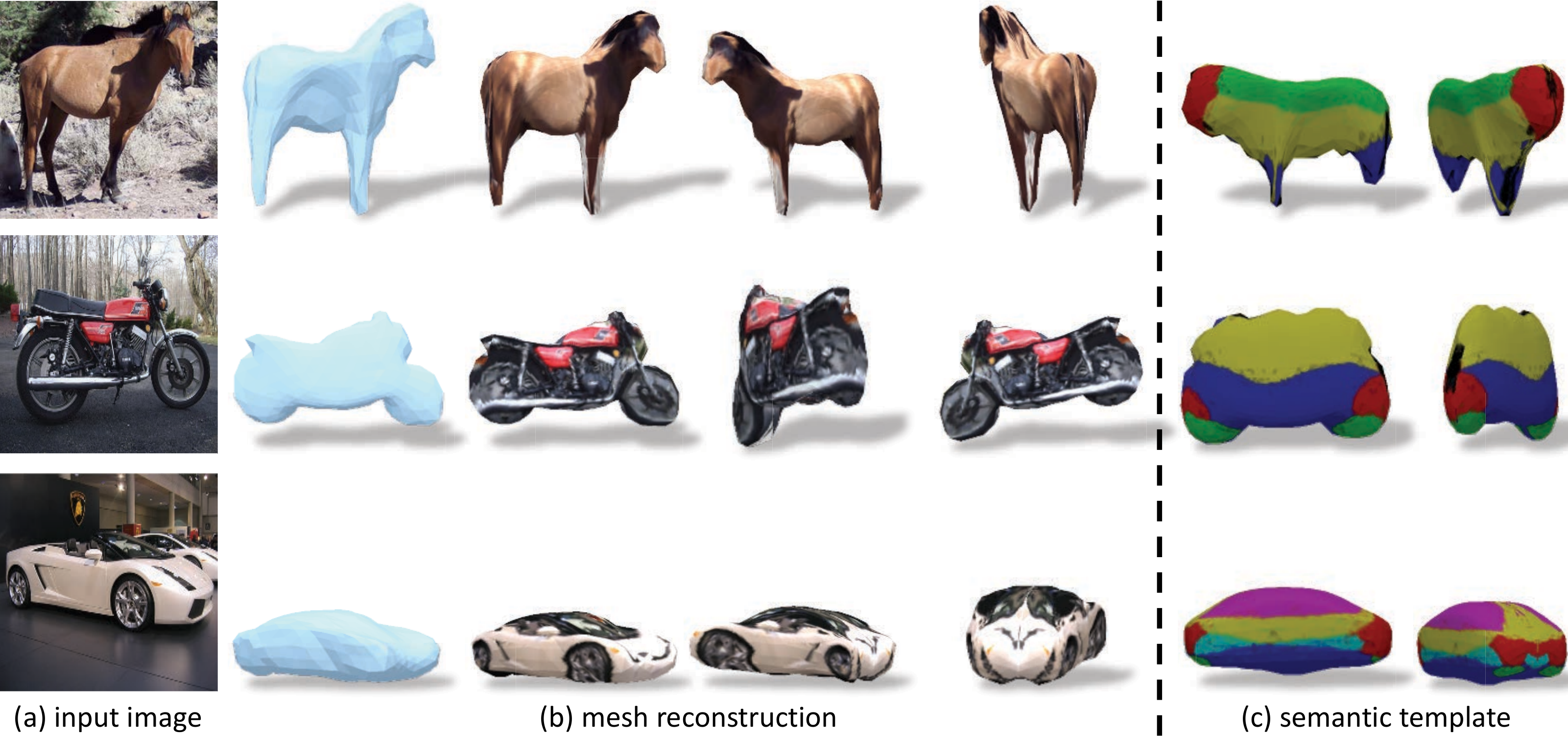}
    {\caption{\small{\textbf{More reconstruction results.} Visualization of instance-level reconstructions and semantic templates for the \textit{horse}, \textit{motorbike} and \textit{car} categories.}}\label{fig:other_results}}
    \vspace{-2mm}
\end{figure}

We show the results of 3D reconstruction from each single-view image in Figure~\ref{fig:exp_cub} (b)-(d) and Figure~\ref{fig:other_results} (b). Our model can reconstruct instances from the same category with highly divergent shapes, e.g., a thin bird in (b), a duck in (c) and a flying bird in (d). 
Our model also correctly maps the texture from each input image onto its 3D mesh, e.g., the eyes of each bird as well as fine textures on the back of the bird.
Furthermore, the renderings of the reconstructed meshes under the predicted camera poses (2nd and 3rd columns in Figure~\ref{fig:exp_cub} and Figure~\ref{fig:other_results}) match well with the input images in the first column, indicating that our model accurately predicts the original camera view.

\vspace{-2mm}
\subsubsection{Improving SCOPS by 3D Reconstruction}
In Figure~\ref{fig:exp_scops}, we visualize the results of improving SCOPS~\cite{hung:CVPR:2019} with our 3D reconstruction network as discussed in Section~\ref{sec:scops}. Thanks to our learned canonical semantic UV map, the improved SCOPS method is able to predict the correct parts and accurately localizes them with a more precise size (head and neck parts in column 1,2,3). 

\vspace{-2mm}
\subsection{Quantitative Evaluations}\vspace{-2mm}
In this section, we quantitatively evaluate the reconstruction network in terms of shape, texture and camera pose prediction of non-rigid (bird~\cite{wah2011caltech}) as well as rigid (car~\cite{xiang2014beyond}) objects. Our model cannot be quantitatively evaluated on other categories (e.g., horse, cows, penguins, etc.) due to a lack of ground truth keypoints, 3D meshes or camera poses in these datasets.
Furthermore, since the ground truth textures and camera poses are also not available for the bird and car categories, we evaluate them through the task of keypoint transfer. 
Given a pair of source and target images of two different object instances from a category, we map a set of annotated keypoints from the source image to the target image by first mapping them onto the learned template and then to the target image. Each mapping can be carried out by either the learned texture flow or the camera pose, as explained below.
%
%

\begin{table*}[t]
\centering
  \caption{\small{Quantitative evaluation of mask IoU and keypoint transfer (KT) on the CUB dataset~\cite{wah2011caltech}. The comparisons are against the baseline supervised models~\cite{cmrKanazawa18,kulkarni2019csm}.}} 
  \label{tab:quanti}
  {\footnotesize{
  \begin{tabular}{c|c|c|c}
  \hline
  (a) Metric & (b) CMR~\cite{cmrKanazawa18} & (c) CSM~\cite{kulkarni2019csm} & (d) Ours \\
  \hline
  Mask IoU~$\uparrow$ & 0.706 &- & 0.734 \\
  KT (Camera)~$\uparrow$ & 47.3 & - & 51.2\\
  KT (Texture Flow)~$\uparrow$ & 28.5 & 48.0 & 58.2\\
  \hline 		 
  \end{tabular}
  }}
  \vspace{-2mm}
\end{table*}

\begin{table*}[t]
\centering
  \caption{\small{Ablation studies of each proposed module by evaluating mask IoU and keypoint transfer (KT) on the CUB-200-2011 dataset~\cite{wah2011caltech}. 
}}
  \label{tab:ablation}
  {\footnotesize{
  \begin{tabular}{c|c|c|c|c}
  \hline
   (a) Metric & (b) Ours & (c) w/o $L_\mathrm{tcyc}$ & (d) w/o $L_\mathrm{sv}$ \& $L_\mathrm{sp}$ & (e) with original~\cite{hung:CVPR:2019}\\
  \hline
  Mask IoU~$\uparrow$ & 0.734 & 0.731 & 0.744 & 0.731\\
  KT (Camera)~$\uparrow$ &  51.2& 48.5& 29.0 & 48.7\\
  KT (Texture Flow)~$\uparrow$ & 58.2& 51.0& 32.8 & 52.9\\
  \hline 		 
  \end{tabular}
  }}
  \vspace{-2mm}
\end{table*}
\label{sec:exp_quanti}

\vspace{-2mm}
\subsubsection{Shape Reconstruction Evaluation}

We first evaluate shape reconstruction on the bird category. Due to a lack of ground truth 3D shapes in the CUB-200-2011 dataset~\cite{wah2011caltech}, we follow~\cite{cmrKanazawa18} and compute the mask reprojection accuracy -- the intersection over union (IoU) between rendered and ground truth silhouettes.
As shown in Table~\ref{tab:quanti}, our model is able to achieve comparable if not better mask reprojection accuracy compared to CMR~\cite{cmrKanazawa18}, which unlike our method is learned with additional supervision from semantic keypoints. This indicates that our model is able to predict 3D mesh reconstructions and camera poses that are well matched to the 2D observations.
%

Next, we evaluate shape reconstruction on the car category. Although PASCAL3D+~\cite{xiang2014beyond} provides ``ground truth'' meshes (the most similar ones to the image in a mesh library), our reconstructed meshes are not aligned with these ``ground truth'' meshes since our self-suerpvised model is free to learn its own ``canonical reference frame''. 
Thus, to quantitatively evaluate the intersection over union (IoU) between the two meshes, following CMR~\cite{cmrKanazawa18}, we exhaustively search a set of scale, translation and rotation parameters that best align to the ``ground truth'' meshes. Our method achieves an IoU (0.62) that is comparable to CMR~\cite{cmrKanazawa18} (0.64), even though the latter is trained with keypoints supervision.

\vspace{-2mm}
\subsubsection{Texture Reconstruction Evaluation via Keypoint Transfer}
Given an annotated keypoint $k^s$ in a source image ($s$), we map it to a triangle face ($F_j$) on the template using its learned flow $I_\mathrm{flow}^s$. We then find all the pixels ($\Omega_j$) in a target image ($t$) that are mapped to the same triangle face $F_j$, by its texture flow $I_\mathrm{flow}^t$ and compute the geometric center of all pixels in $\Omega_j$.  
We compare the location of the geometric center of $\Omega_j$ to the ground truth keypoint $k^t$ and find the percentage of correct keypoints (PCK) as those that fall within a threshold distance $\alpha=0.1$ of each other~\cite{kulkarni2019csm}. 
%
%
Figure~\ref{fig:kp_transfer} (a) demonstrates qualitative visualization of the keypoint transfer results using texture flows and Table~\ref{tab:quanti} shows that the texture flow learned by our method, even without supervision, outperforms the 2D$\rightarrow$3D mappings learned by the supervised methods~\cite{cmrKanazawa18, kulkarni2019csm}.


\vspace{-4mm}
\subsubsection{Camera Pose prediction Evaluation via Keypoint Transfer}
To find the 3D template's vertex $v$ that corresponds to an annotated 2D keypoint $k^s$ of a source image, we first render all 3D vertices using the source image's predicted pose $\theta^s$. 
Then, $v$ is the vertex whose 2D projection lies closest to the keypoint $k^s$.
Next, we render the point $v$ with the target image's predicted pose $\theta^t$ and compare it to its ground truth keypoint $k^t$ to compute PCK.
%
%
%
%
Figure~\ref{fig:kp_transfer} (b) demonstrates the keypoint transfer results by predicted camera pose. Table~\ref{tab:quanti} shows that our model achieves favourable performance against the baseline method~\cite{cmrKanazawa18}. 


\vspace{-2mm}
\subsection{Ablation Studies}
\vspace{-2mm}
\label{sec:exp_ablation}
In this section, we discuss the contribution of each proposed module: (i) The semantic consistency constraint discussed in Section~\ref{sec:semantics}. (ii) The texture cycle consistency introduced in Section~\ref{sec:tcyc}. (iii) The improved SCOPS method introduced in Section~\ref{sec:scops}. We evaluate on the CUB-200-2011 dataset~\cite{wah2011caltech} and use the mask reprojection accuracy as well as the keypoint transfer (via texture flow and via camera pose) accuracy discussed in Section~\ref{sec:exp_quanti} as our metrics.

\vspace{-2mm}
\subsubsection{The Semantic Consistency Constraint}
%
As shown in Table~\ref{tab:ablation} (b) \textit{vs.} (d)
our baseline model trained without semantic consistency constraint performs much worse at the keypoint transfer task than our full model, indicating this baseline model predicts incorrect texture flow and camera views. 
We note that this baseline model achieves better mask IoU because the model trained without any constraint is more prone to overfit to the 2D silhouette observations.

\vspace{-2mm}
\subsubsection{The Texture Cycle Consistency} Our model trained without the texture cycle consistency constraint achieves worse performance (Table~\ref{tab:ablation} (b) \textit{vs.}(c)) at transferring keypoints using the predicted texture flow. This proves the effectiveness of the texture cycle consistency constraint in encouraging the model to learn better texture flow.

\vspace{-2mm}
\subsubsection{SCOPS: Before \& After Improvement} 
\begin{figure}[t]
\centering
\begin{minipage}{.48\textwidth}
  \centering
  \includegraphics[height=.7\linewidth]{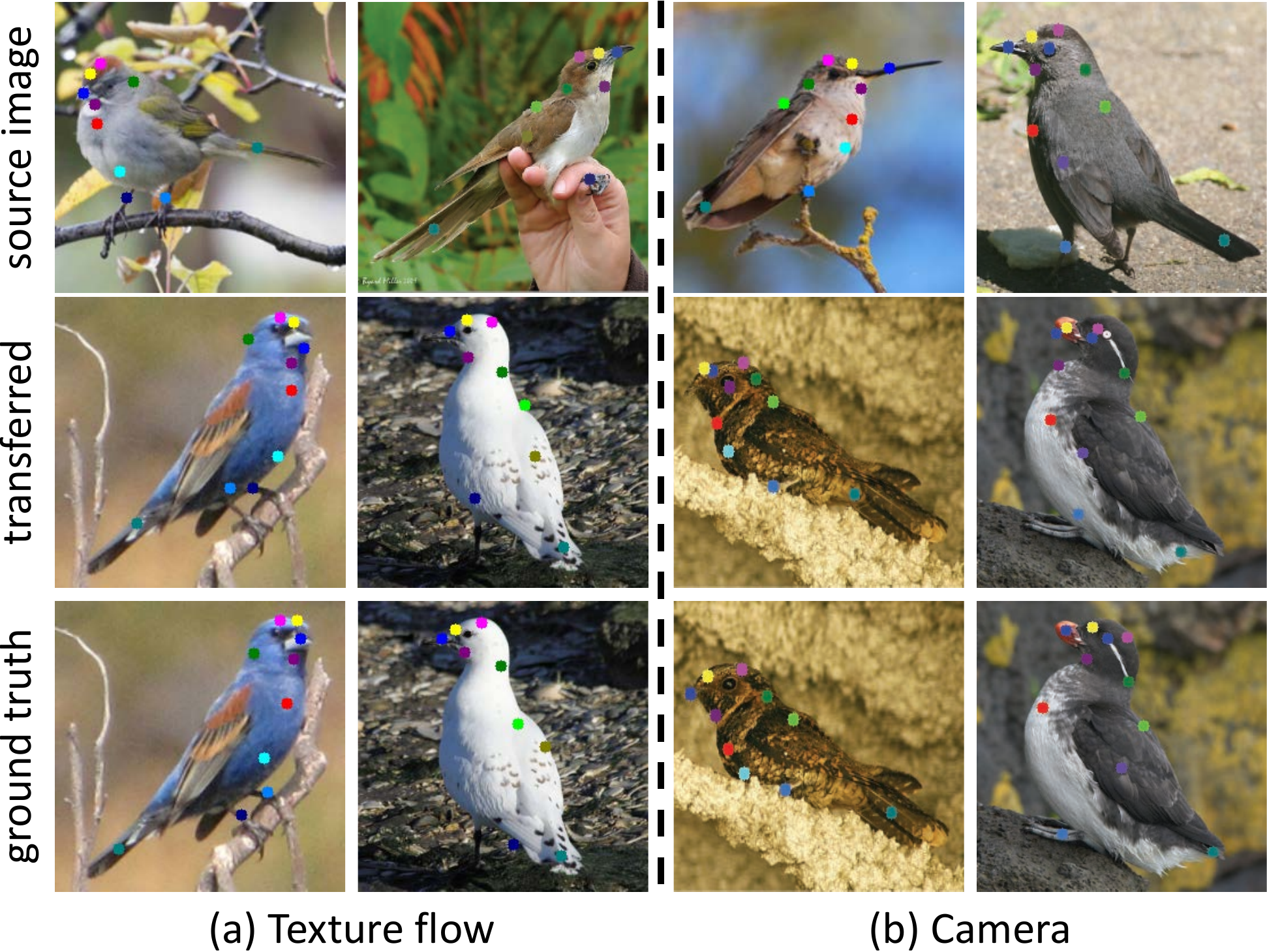}
  \caption{\small{ \textbf{Qualitative visualization of keypoint transfer.} We show comparison with the ground truth keypoints in each column.} }
  \label{fig:kp_transfer}
\end{minipage}
\hfill
\begin{minipage}{.48\textwidth}
  \centering
  \includegraphics[height=.7\linewidth]{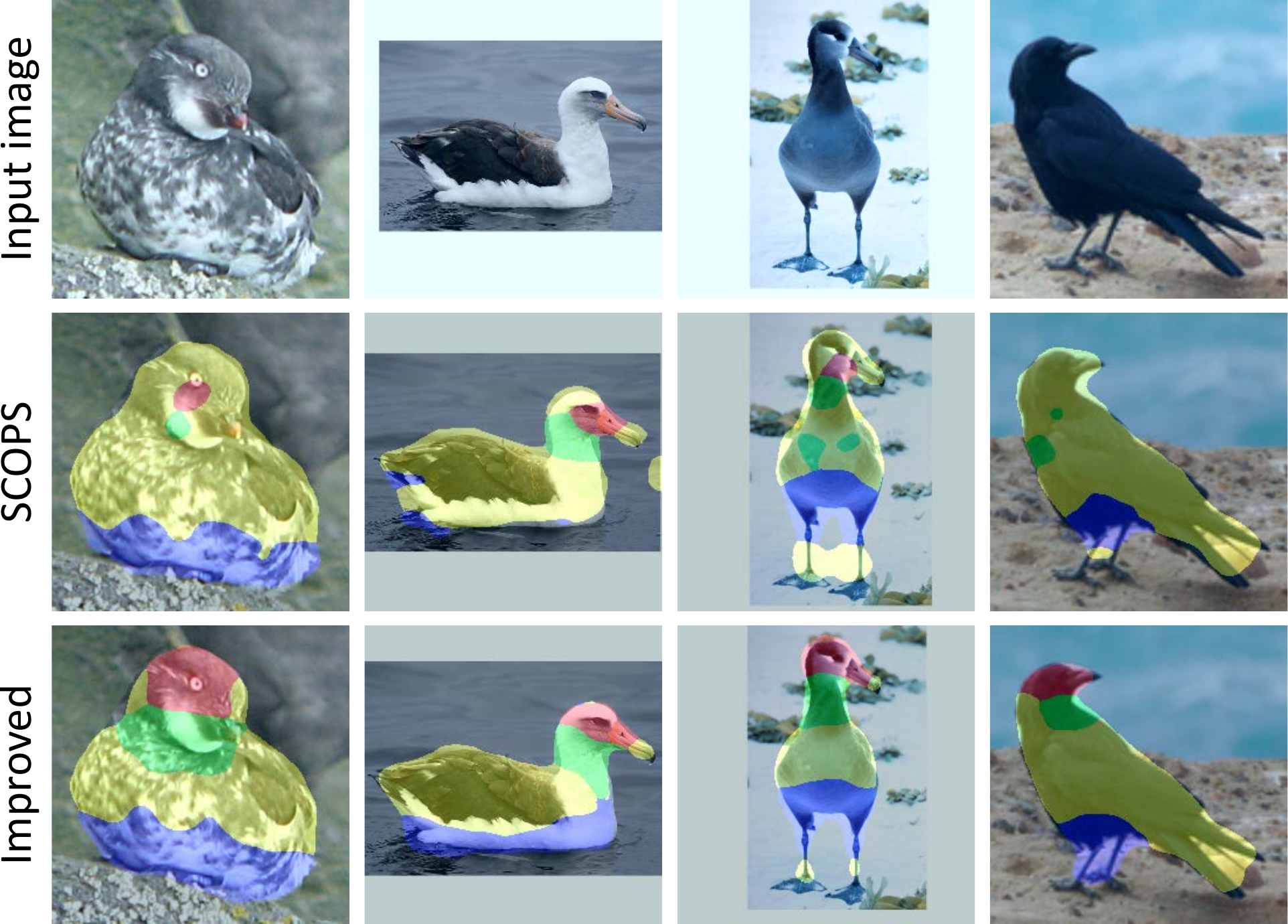}
  \caption{\small{ \textbf{Improvements to the SCOPS~\cite{hung:CVPR:2019} method.} Notice the more consistent size and shape of part segments with the improved method.}}
  \label{fig:exp_scops}
\end{minipage}%
\vspace{-2mm}
\end{figure}

Our method improves SCOPS by segmenting parts more consistently in terms of their shape and size as shown in Figure~\ref{fig:exp_scops}. However, this is non-trivial to quantify numerically as the ground-truth segmentation labels for the parts are not available in all the dataset that we use~\cite{wah2011caltech,xiang2014beyond,deng2009imagenet,kuznetsova2018open}. Instead, we indirectly measure the improvement by training two models, each of which uses the semantic part segmentation predicted either by the original or the improved SCOPS method. As shown in Table~\ref{tab:ablation} (b) \textit{vs.}(e), our keypoint transfer performance drops by 5.3\% and 2.5\% via texture flow and camera pose if we use the original SCOPS model. More qualitative visualizations of the improvement of SCOPS can be found in the appendix.

\vspace{-2mm}
\section{Conclusion}
\vspace{-2mm}
In this work, we learn a model to reconstruct 3D shape, texture and camera pose from single-view images, with only a category-specific collection of images and silhouettes as supervision.
The self-supervised framework enforces semantic consistency between the reconstructed meshes and images and largely reduces ambiguities in the joint prediction of 3D shape and camera pose from 2D observations.
It also creates a category-level template and a canonical semantic UV map, which capture the most representative shape characteristics and semantic parts of objects in each category, respectively.
Experimental results demonstrate the efficacy of our proposed method in comparison to the state-of-the-art supervised category-specific reconstruction methods.

\chapter*{Appendix}


In this appendix, we provide additional details, discussions, and experiments to support the original submission. 
We first discuss implementation details in Section~\ref{sec:imple_detail}. Then, we visualize the contribution of each module via ablation studies in Section~\ref{sec:ablation}. We further present more quantitative and qualitative results in Section~\ref{sec:quantitative} and Section~\ref{sec:qualitative}. Finally, we describe failure cases and limitations of the proposed method in Section~\ref{sec:failure}. 

\begin{figure*}[h]
\centering
\includegraphics[width=\linewidth]{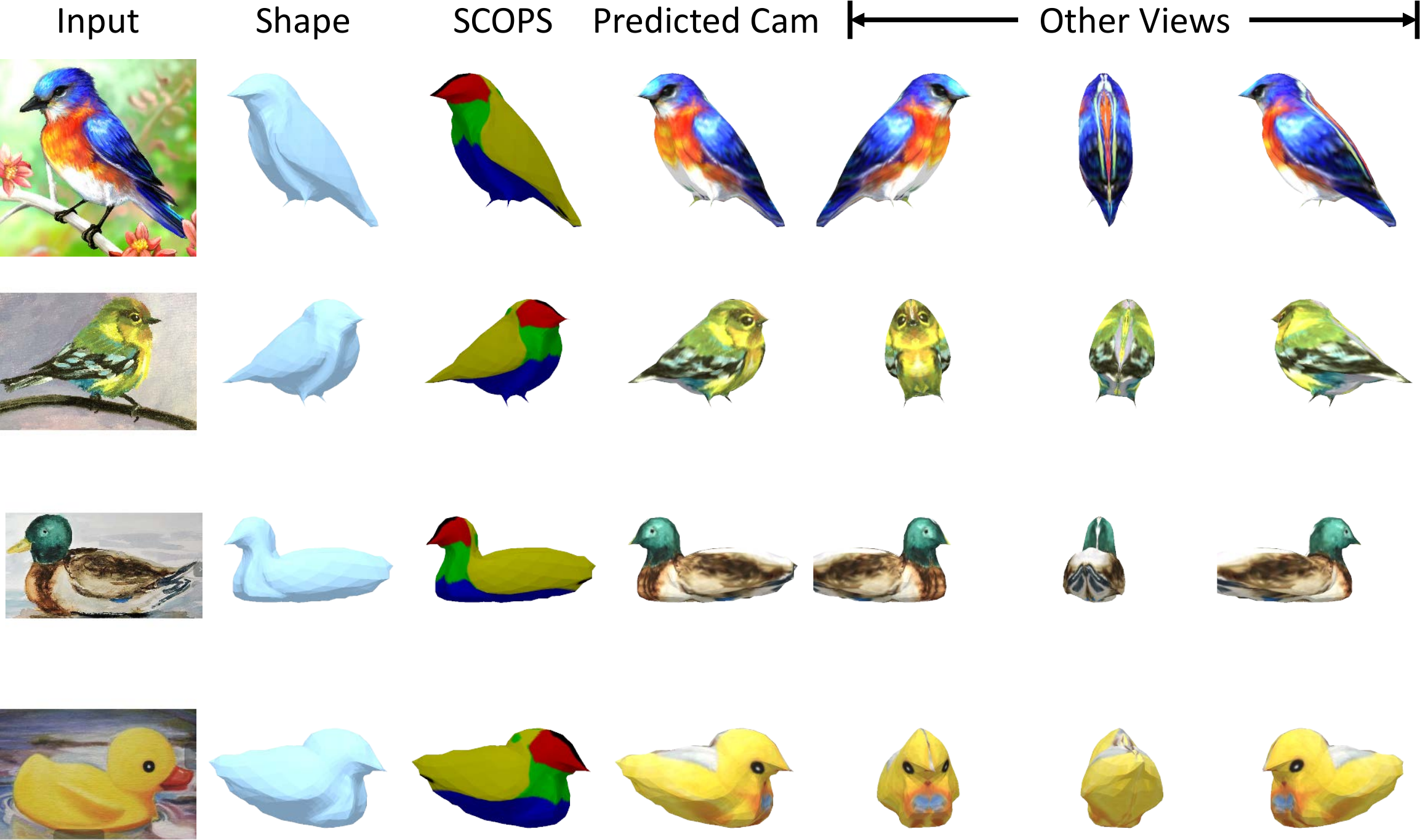}
\caption{Results of applying our reconstruction model on bird paintings.}\vspace{-3mm}\label{fig:bird_paintings}
\end{figure*}

\section{Implementation Details}
\label{sec:imple_detail}
\subsection{Selective Aggregation in the M-step}
\subsubsection{Computing Category-level Template}
In the M-step (Section 3.2 in the submission), we update the template by decoding the averaged shape feature via the shape decoder. Instead of using all training samples to obtain the averaged feature, we select a subset of the training samples to form a set $\mathcal{Q}$ and compute the averaged feature for the samples in this set. In the following, we explain why and how to form this set $\mathcal{Q}$ used in Eq.(4) of the submission. 
Empirically we found that for several categories, there exist ambiguities that produce inconsistent mesh reconstructions, e.g., side-view images of horses could be reconstructed with their heads on either the left or the right side. 
Aggregating such instance meshes leads to incorrect estimation of the category-level template.  
%
To resolve this, we select a subset of reconstructed meshes whose viewpoints roughly match (e.g. horses with heads on the left side). To do so, from the meshes reconstructed for all the training images, we first choose the instance with the most ``reliable'' reconstruction results, i.e., the instance whose rendered silhouette has the largest intersection over union (IoU) with its corresponding ground truth silhouette, as an exemplar (e.g. a horse shape with its head on the left).
We then use the top $k$ training samples with meshes that are most similar to the exemplar mesh to form the subset $\mathcal{Q}$ in Eq.(4) (e.g., all chosen horse samples have heads on the left). We measure the similarity between an individual instance mesh and the exemplar mesh by computing the IoU between their rendered silhouettes. 

\subsubsection{Computing Canonical Semantic UV Map}
Similarly, when we update the canonical semantic UV map using Eq.(1) (see Section 3.2 in the submission), to avoid using training samples with outliers, e.g., those caused by inaccurate prediction of $I_\mathrm{flow}^i$, we choose an exemplar training example with the smallest perceptual distance objective (see Section 3.2 in the submission), and form the set $\mathcal{U}$ of the top $k$ training samples that have the most similar semantic UV maps (as measured by the L2 norm) to the exemplar.

\subsection{Network Architecture and Other Objectives}
\subsubsection{Network Architecture} 
\begin{figure}[t]
    \centering
    \includegraphics[width=\linewidth]{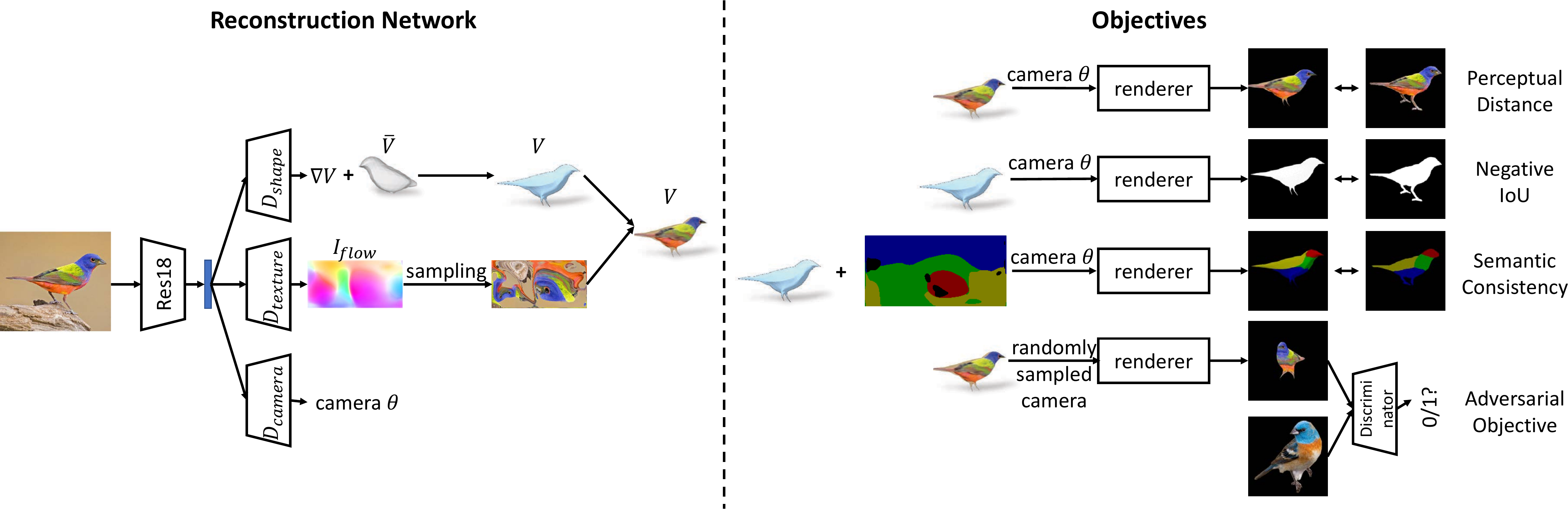}
    {\caption{\footnotesize Network Architecture and Objectives. }\label{fig:architecture}}
\end{figure}

We present the details of our network architecture as well as training objectives in Figure~\ref{fig:architecture}. 
We use the same network as in CMR~\cite{cmrKanazawa18}, where: (i) the encoder is the ResNet18 network~\cite{he2016deep} with four residual blocks and is pretrained on the ImageNet~\cite{deng2009imagenet} dataset, (ii) the shape decoder consists of one fully connected layer to decode shape deformation $\Delta V$, (iii) the texture decoder contains two fully-connected layers followed by eleven upsample and convolution layers to predict the texture flow $I_\mathrm{flow}$, (iv) the camera pose decoder contains three parallel fully connected layers to predict the scale, translation and rotation, respectively and these three parameters together compose the camera pose $\theta$.  Note that we use the one camera hypothesis in the first EM training round and use the multiple camera hypothesis (eight camera hypothesis) as in~\cite{kulkarni2019csm,insafutdinov18pointclouds,mvcTulsiani18} to avoid local minima in the subsequent rounds. 
To render the reconstructed meshes, we utilize the Soft Rasterizer~\cite{liu2019softras} instead of the Neural Mesh Renderer~\cite{kato2018renderer} used in the CMR~\cite{cmrKanazawa18}.
This is because it provides the probability map described in Section 3.3 for the texture cycle consistency constraint.

\subsubsection{Smoothness Term}
In addition to the objectives discussed in Section 3.2 of the submission, we further utilize a graph Laplacian constraint to encourage the reconstructed mesh surface to be smooth~\cite{cmrKanazawa18,liu2019softras}, and adopt an edge regularization to penalize irregularly-sized faces as in~\cite{wang2018pixel2mesh,gkioxari2019mesh}. More details can be found in ~\cite{cmrKanazawa18,liu2019softras,wang2018pixel2mesh,gkioxari2019mesh}.

\subsubsection{Adversarial Training}
To constrain the reconstructed meshes to look plausible from all views, we also introduce adversarial training~\cite{goodfellow2014generative} into the mesh reconstruction framework~\cite{kato2019vpl}.
We render the reconstructed mesh from a randomly sampled camera pose to obtain an image ${I_\mathrm{rd}}$, and pass it together with a random real image $I_\mathrm{rl}$ into a discriminator. 
By learning to discriminate between the real and rendered images, the discriminator learns shape priors and constrains the reconstruction model to generate meshes that are plausible from all viewpoints. 
The adversarial loss is:
\begingroup\small
\begin{eqnarray}\vspace{-2mm}
 \label{eq:advloss}
 \begin{aligned}
     L_\mathrm{adv}(R,D) = {\mathop{\mathbb{E}}}_{I_\mathrm{rl}}[\log{D(I_\mathrm{rl})}] + {\mathop{\mathbb{E}}}_{I_\mathrm{rd}}[\log{(1-D(I_\mathrm{rd}))}],
 \end{aligned}
 \end{eqnarray}\endgroup
where $R$ and $D$ are the reconstruction and discriminator networks, respectively. Figure~\ref{fig:architecture} illustrates the adversarial objective.

\subsection{Network Training}
\begin{figure}[h]
    \centering
    \includegraphics[width=0.9\linewidth]{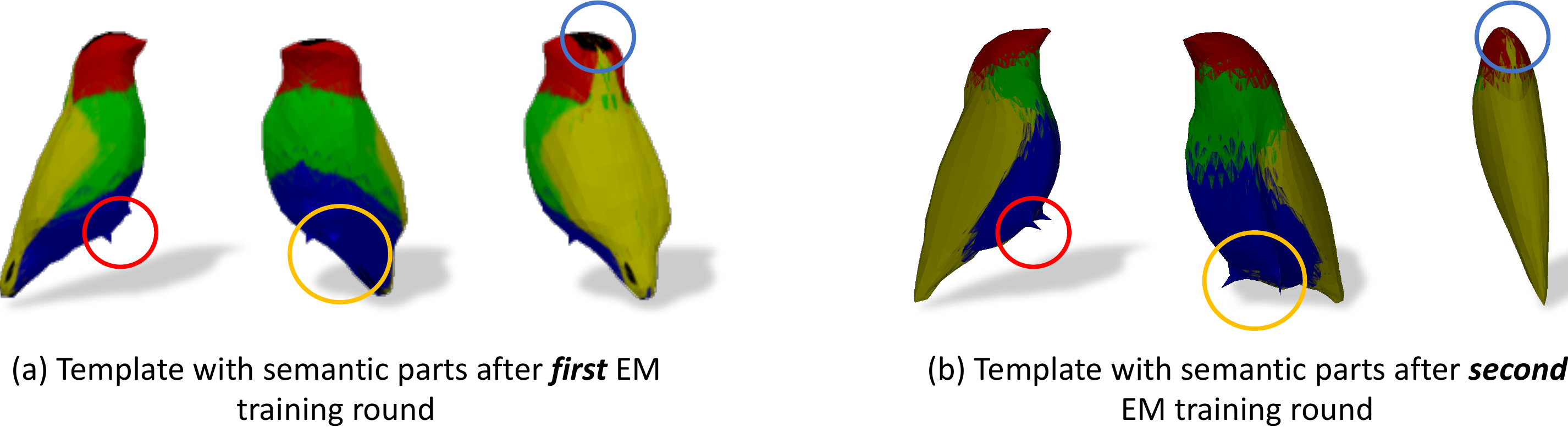}
    {\caption{\footnotesize Visualization of the learned template and semantic parts. Notice the improvements of the template after the second M-step compared to the first, i.e., better feet shape in the red and yellow circles, and a part of the head (blue circle) that was mistakenly assigned to the background (colored in black) in the first step is corrected (colored in red) in the second M-step.}\label{fig:semantic_templates}}
    \vspace{-4mm}
\end{figure}

We train the reconstruction network with an initial learning rate of $1e-4$ and gradually decay it by a factor of $0.5$ every 2000 iterations. The network is trained for two EM training rounds (each training round contains one E and M-step) on four NVIDIA Tesla V100 GPUs for two days. 
We found that two rounds of EM training are sufficient to generate high-quality reconstruction results.
During the inference stage, the model takes 0.022 seconds to reconstruct a 3D mesh from a $256\times256$ sized single-view image on a single NVIDIA Tesla V100 GPU. 
In Figure~\ref{fig:semantic_templates}, we show the learned template shape as well as the semantic parts after the first (left figure) and second M-steps (right figure), where both the template shape and the semantic parts after the second M-step are better than the first.


\section{Ablation Studies}
\label{sec:ablation}
\subsection{Ablation Studies for Different Objectives}
\label{sec:ablation_objective}
\begin{table}[t]
\centering
  \caption{\small{Settings of each baseline models in Section~\ref{sec:ablation_objective}.}}
  \label{tab:ablation_setting}
  {\footnotesize 
  \begin{tabular}{c|c|c|c}
  \hline
  Module & category-level template & semantic consistency & adversarial training\\
  \hline
  baseline (a) & $\times$ & $\times$ & $\times$\\
  baseline (b) & \checkmark & $\times$ & $\times$\\
  baseline (c) & \checkmark &  \checkmark & $\times$\\
  \textbf{Ours} & \checkmark & \checkmark & \checkmark\\
  \hline 		 
  \end{tabular}\vspace{-2mm}
  }
\end{table}

\begin{figure*}[t]
\centering
\includegraphics[width=0.8\linewidth]{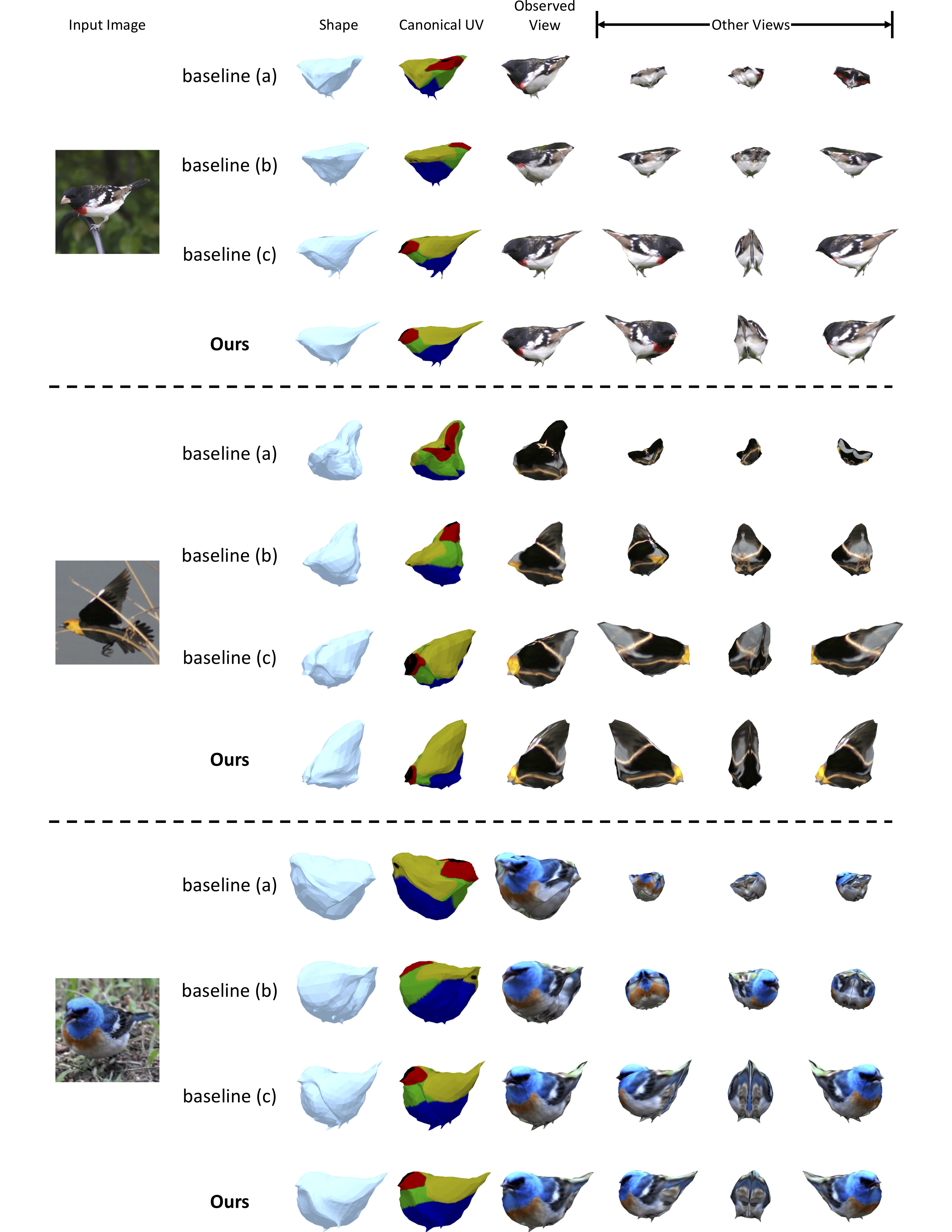}
\caption{Visualization of the contribution of each module. The settings of baselines (a), (b), (c) can be found in Table~\ref{tab:ablation_setting}}\label{fig:ablation}
\end{figure*}

We show the results of three baselines in Figure~\ref{fig:ablation}. The experimental settings for each are illustrated in Table~\ref{tab:ablation_setting} and are the following: (a) a basic model trained with only the texture cycle consistency constraint described in Section 3.3 in the submission, but without any other proposed modules, i.e., the category-level template, the semantic consistency constraint and the adversarial training; (b) learning the model in (a) together with the category-level template; and (c) learning the model in (b) with the additional semantic consistency constraint.



As shown in Figure~\ref{fig:ablation}, the basic model (a) reconstructs meshes that only appear plausible from the observed view to match the 2D supervision (images and silhouettes). 
It fails to generate plausible results for unobserved views, e.g., for all the 3 examples. 
On adding template shape learning (Section 3.2 in the submission) to (a), the model in (b) learns more plausible reconstruction results across different views. 
This is because it is easier for the model to learn residuals w.r.t a category-level template compared to w.r.t a sphere, to match the 2D observations. 
However, without semantic part information, the model still suffers from the ``camera-shape ambiguity'' discussed in Section 1 of the manuscript. For instance, the head of the template is deformed to form the tail and the wing's tip in the first and second examples, respectively in Figure~\ref{fig:ablation}.
By additionally including the semantic consistency constraint in the model (c), the network is able to reduce the ``camera-shape ambiguity'' and predict the correct camera pose as well as the correct shape.
Furthermore, adding adversarial training introduces better reconstruction details, as shown in Figure~\ref{fig:ablation} (d). For instance, the bird may have more than two feet without the adversarial training constraint as demonstrated in the third example in Figure~\ref{fig:ablation}.

\begin{figure*}[t]
\centering
\includegraphics[width=0.88\linewidth]{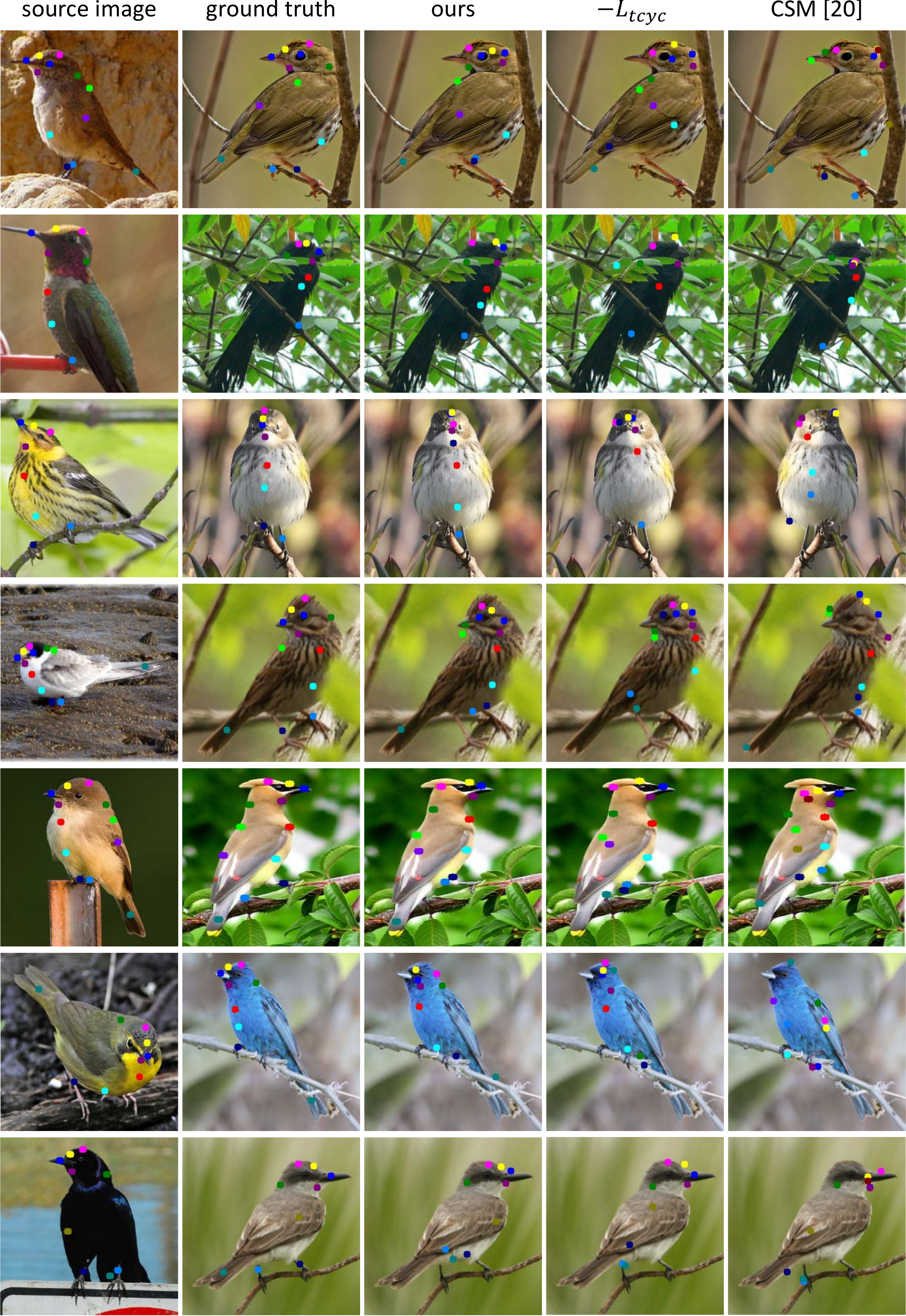}
\caption{Visualization of keypoint transfer using texture flow.}\label{fig:kp_transfer_supp}
\end{figure*}

In addition, we demonstrate the effectiveness of the texture flow consistency constraint by visualizing the keypoint transfer results in Figure~\ref{fig:kp_transfer_supp}. The model trained without this constraint performs worse than our full model, especially when the bird has a uniform color, e.g., the second and the last examples in Figure~\ref{fig:kp_transfer_supp}.
Figure~\ref{fig:kp_transfer_supp} also shows that the proposed method performs favourably against the baseline CSM~\cite{kulkarni2019csm} method.

\begin{figure}[h]
\centering
\begin{minipage}{.48\textwidth}
  \centering
  \includegraphics[height=.5\linewidth]{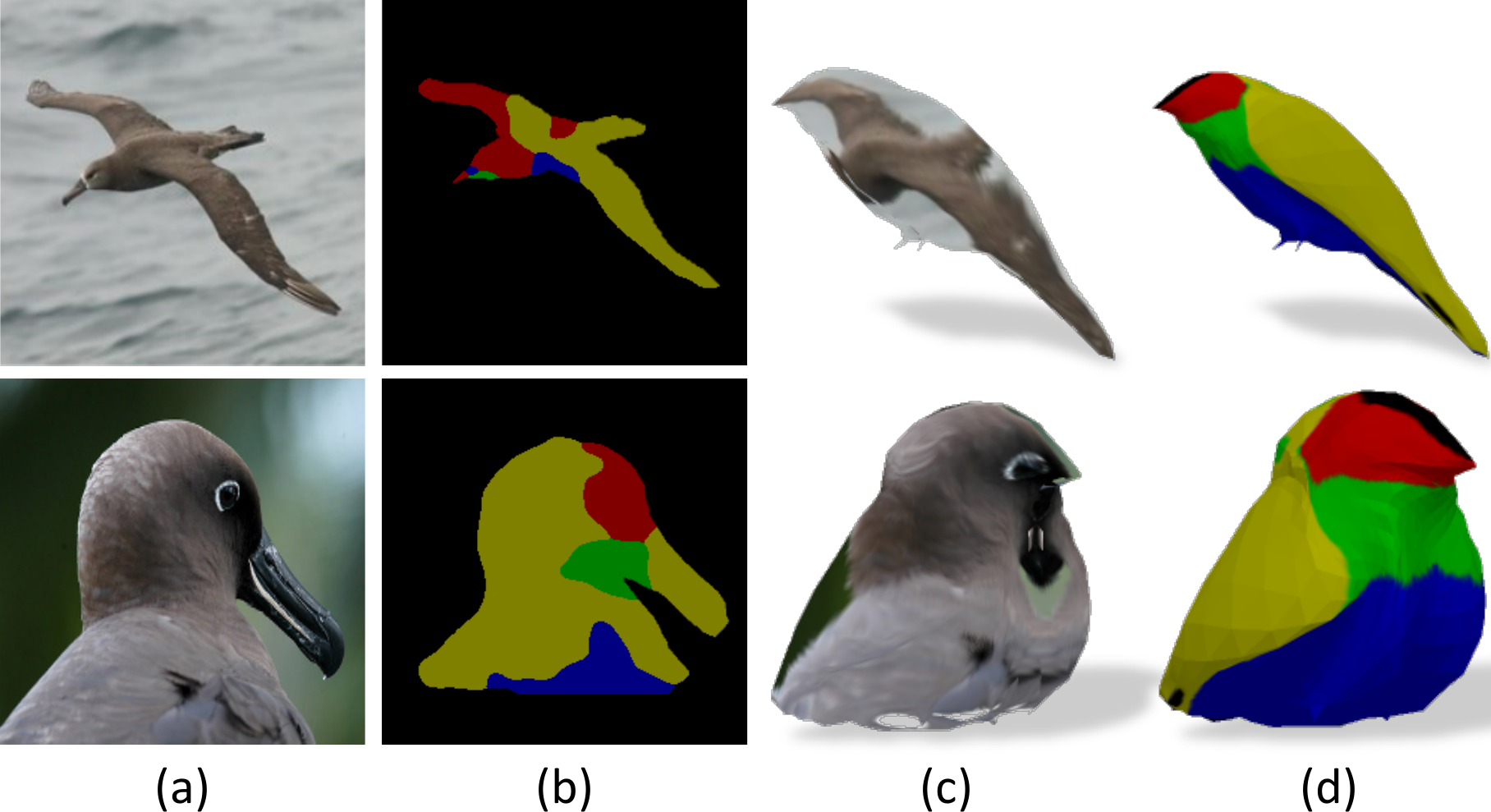}
  {\caption{\footnotesize Failure cases. (a) Input images. (b) Semantic part segmentations predicted by the SCOPS method. (c) Reconstructed meshes. (d) Reconstructed meshes with the canonical semantic UV map.}\label{fig:failure_case}}
\end{minipage}
\hfill
\begin{minipage}{.48\textwidth}
  \centering
  \includegraphics[height=.5\linewidth]{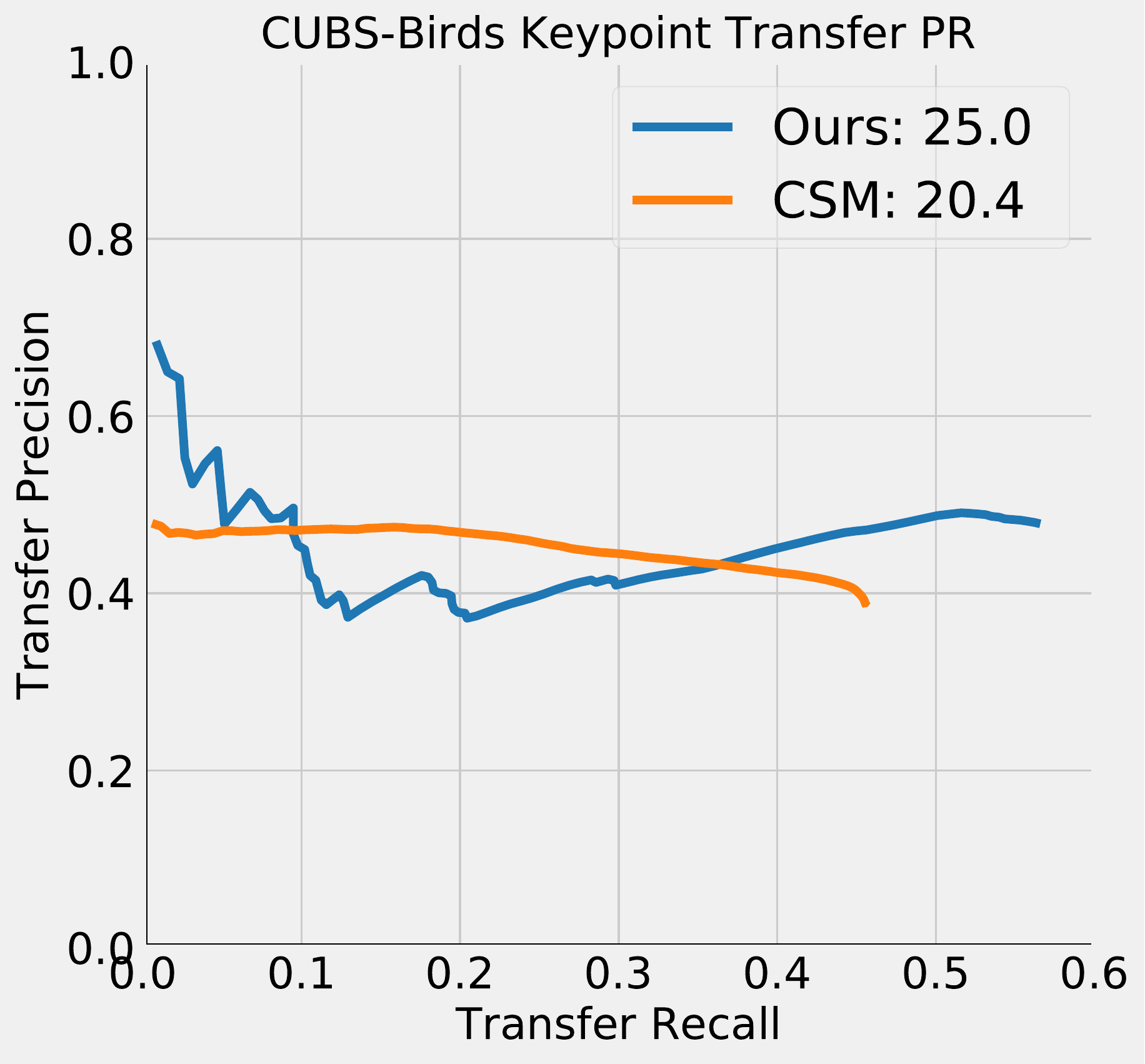}
  \caption{\footnotesize{Keypoint Transfer PR Curves.} The legend of the plot represents the area under the curve, our method achieves an APK of 25.0, which is better than the baseline method~\cite{kulkarni2019csm}.}
  \label{fig:pr}
\end{minipage}%
\vspace{-6mm}
\end{figure}

\subsection{Ablation Studies on Semantic Consistency Constraints}
\begin{table*}[t]
\centering
  \caption{\small{Ablation studies of the probability and vertex-based semantic consistency constraints by evaluating the mask IoU and the keypoint transfer (KT) task on the CUB-200-2011 dataset~\cite{wah2011caltech}. 
}}
  \label{tab:semantic_ablation}
  {\footnotesize{
  \begin{tabular}{c|c|c|c}
  \hline
   (a) Metric & (b) Ours & (c) w/o $L_\mathrm{sv}$ & (d) w/o  $L_\mathrm{sp}$ original~\cite{hung:CVPR:2019}\\
  \hline
  Mask IoU~$\uparrow$ & 0.734 & 0.6069 & 0.6418\\
  KT (Camera)~$\uparrow$ &  51.2&  30.7 & 51.0\\
  KT (Texture Flow)~$\uparrow$ & 58.2& 29.5 & 53.3\\
  \hline 		 
  \end{tabular}
  }}
  \vspace{-2mm}
\end{table*}

We show an ablation study of the probability and vertex-based semantic consistency constraints in Table~\ref{tab:semantic_ablation}, where both constraints contribute to the reconstruction network.

\section{More Quantitative Evaluations}
\label{sec:quantitative}
\subsection{APK Evaluation}
For the keypoint transfer task, in Figure~\ref{fig:pr}, We demonstrate the precision versus recall curve of our method (via texture flow) and of the CSM~\cite{kulkarni2019csm} method on the CUB-200-2011~\cite{wah2011caltech} \textit{test} dataset. Our method, even without the template prior, outperforms the baseline CSM~\cite{kulkarni2019csm} method in terms of the Keypoint Transfer AP metric (APK, $\alpha = 0.1$).

\section{More Qualitative Evaluations}
\label{sec:qualitative}
\begin{figure*}[t]
\centering
\includegraphics[width=0.95\linewidth]{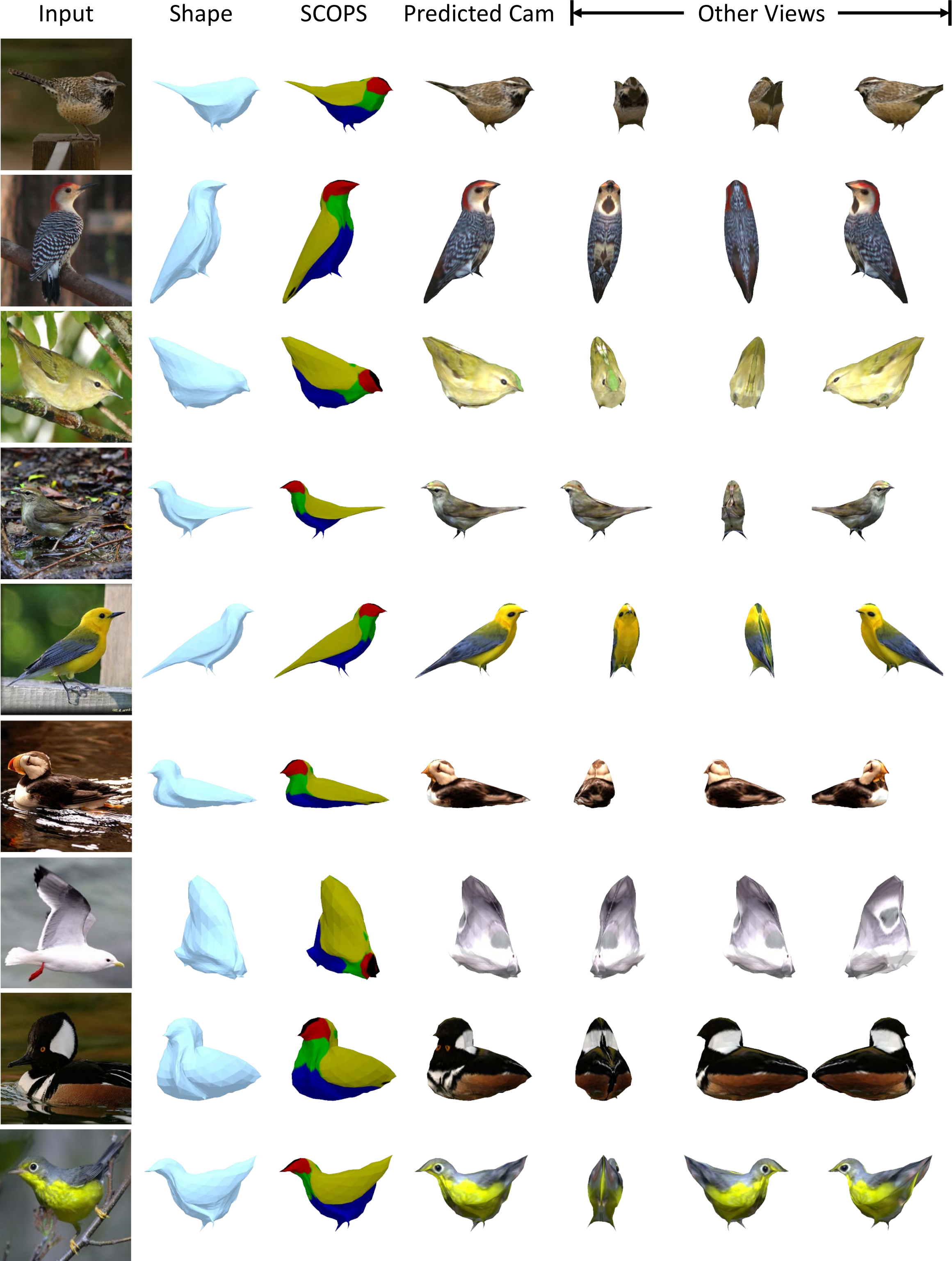}
\caption{More qualitative results of birds.}\label{fig:bird}
\end{figure*}

\begin{figure*}[t]
\centering
\includegraphics[width=0.95\linewidth]{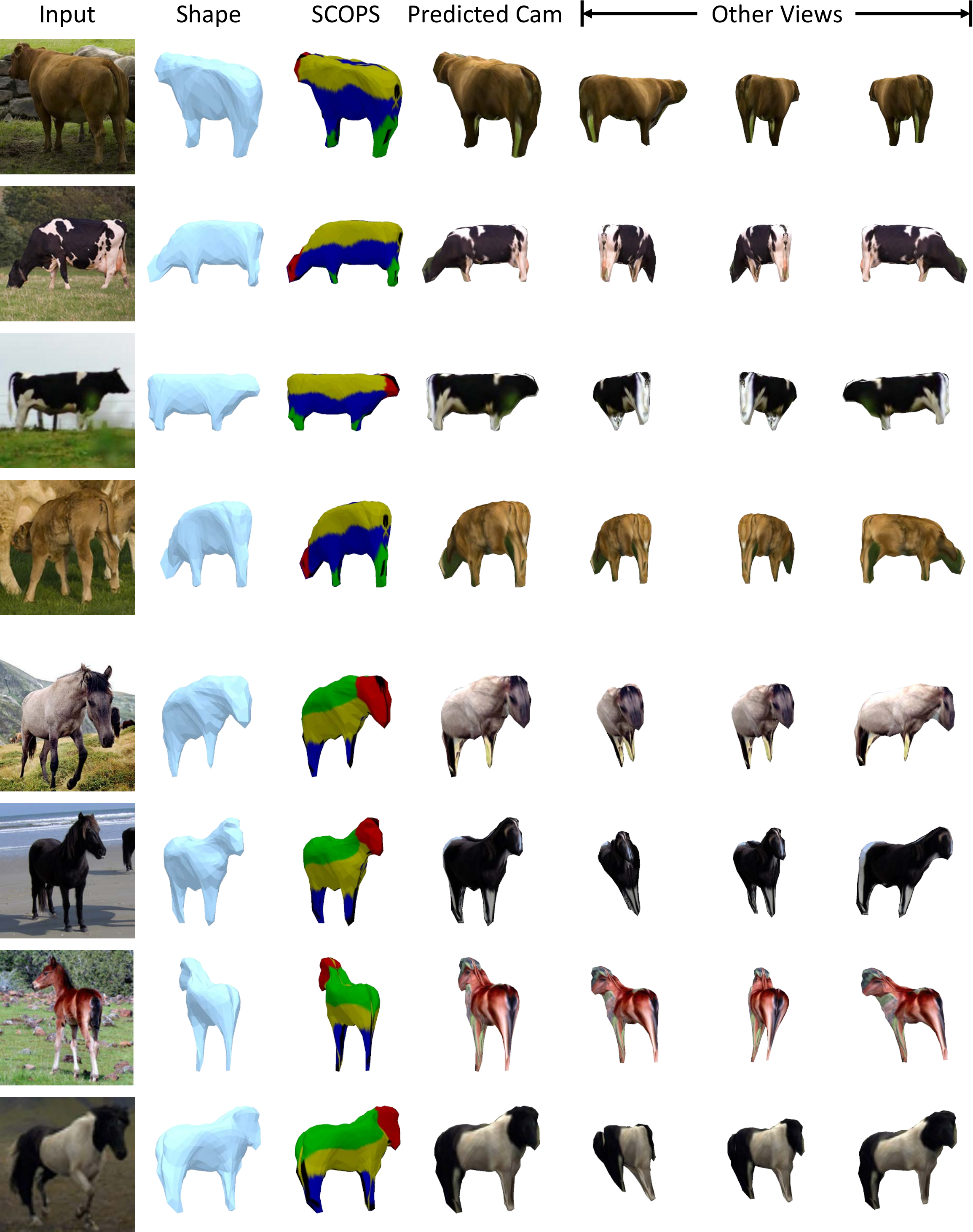}
\caption{More qualitative results of horses and cows.}\label{fig:horse_cow}
\end{figure*}

\begin{figure*}[t]
\centering
\includegraphics[width=\linewidth]{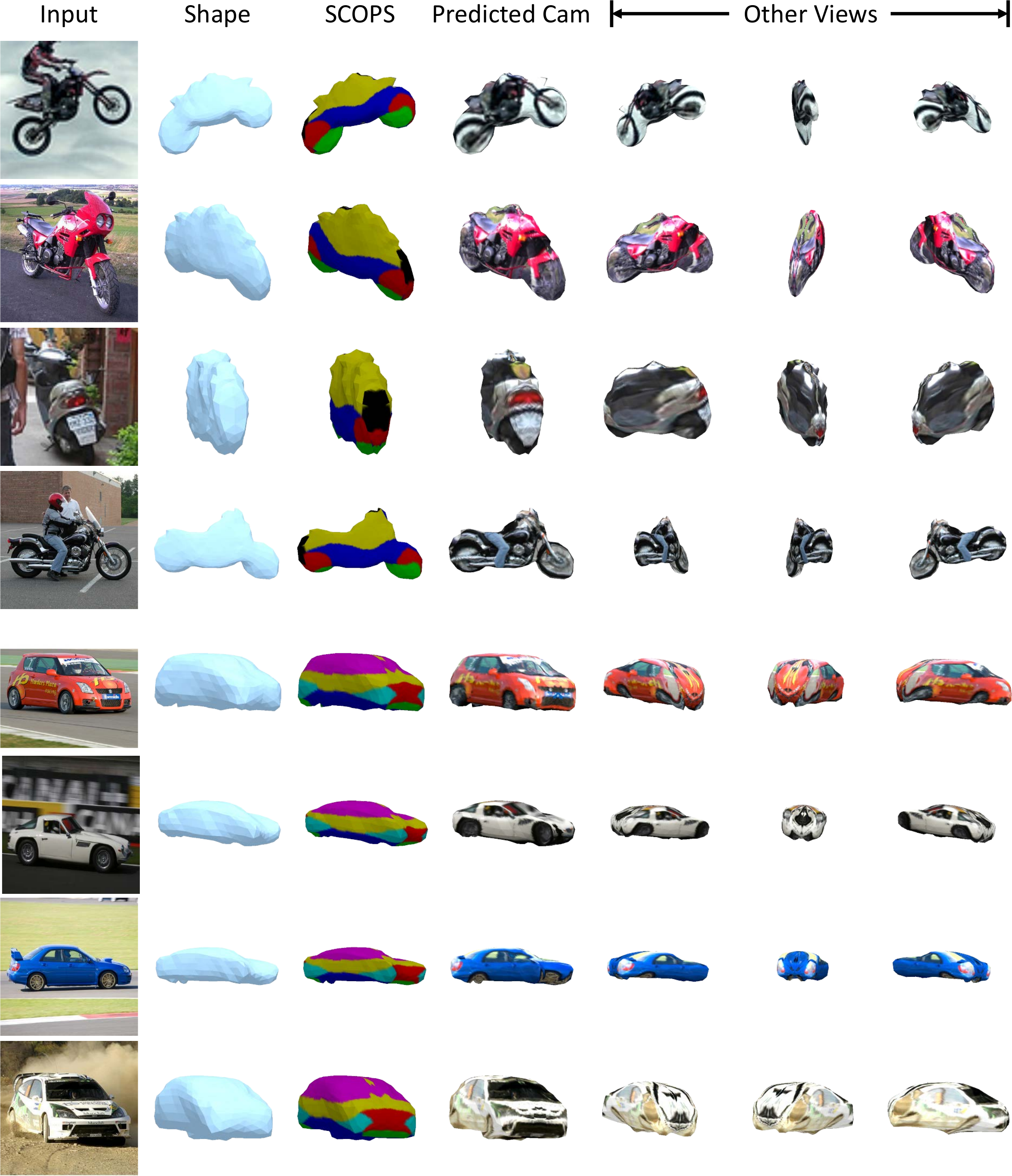}
\caption{More qualitative results of motorbikes and cars.}\label{fig:rigid}
\end{figure*}

\begin{figure*}[t]
\centering
\includegraphics[width=\linewidth]{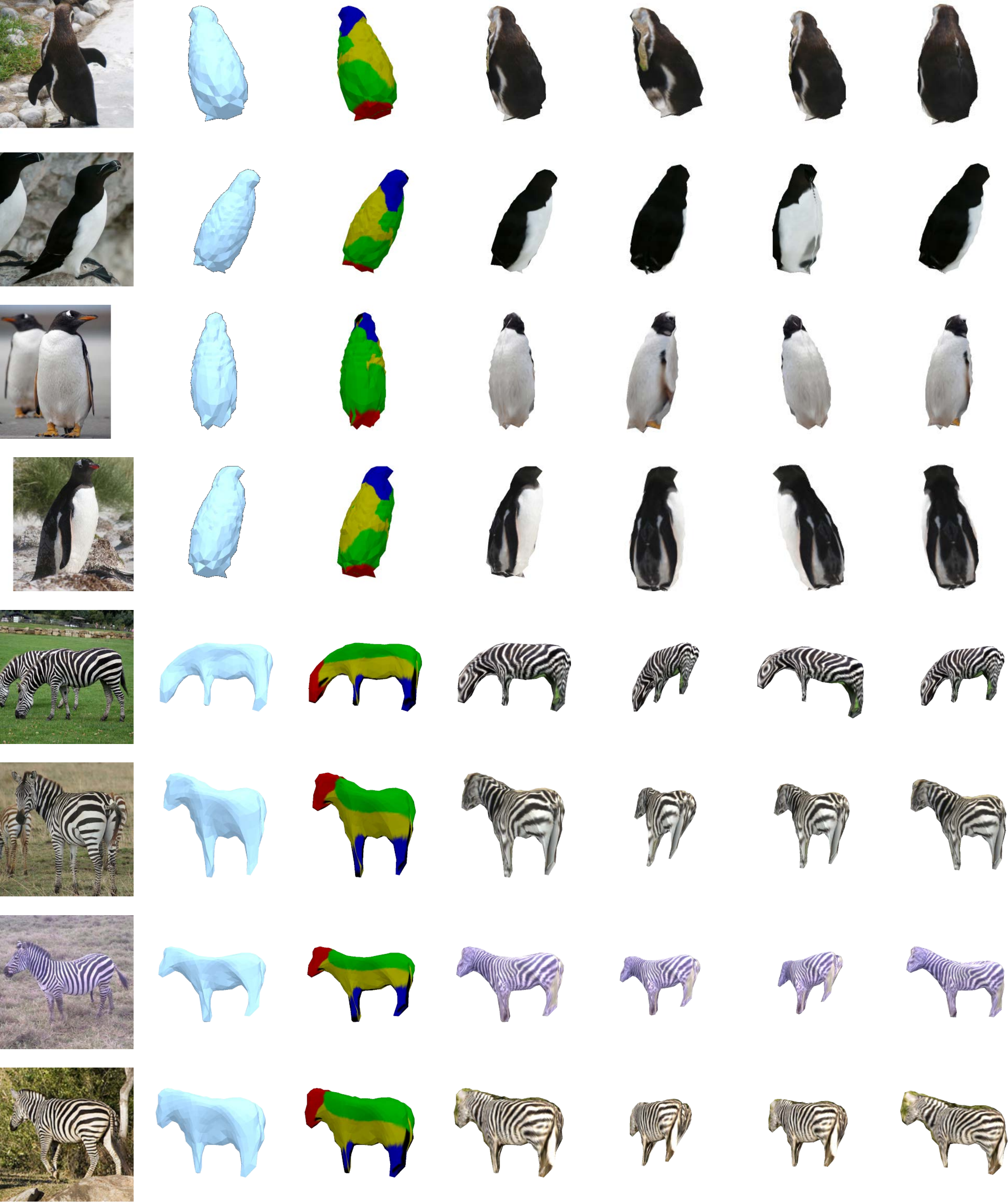}
\caption{More qualitative results of zebras and penguins.}\label{fig:zebra}
\end{figure*}

We show more qualitative results for birds in Figure~\ref{fig:bird}. We also show one application of our model to reconstruct 3D meshes of 2D bird paintings in Figure~\ref{fig:bird_paintings}.
Reconstruction of rigid objects (cars and motorbikes) is demonstrated in Figure~\ref{fig:rigid}, horses and cows in Figure~\ref{fig:horse_cow}, and penguins and zebras in Figure~\ref{fig:zebra}. 
Note that we use six semantic parts for the car category to encourage the SCOPS method~\cite{hung:CVPR:2019} to differentiate between the fronts and the sides of cars. For other objects, we use four semantic parts.

\section{Failure Case and Limitations}
\label{sec:failure}

Our work is the first to tackle the challenging self-supervised single-view 3D reconstruction task. Impressive as the results are, this challenging task is far from being solved. In this section, we discuss typical failure cases and limitations of the proposed method.

First, our method utilizes the SCOPS method to provide semantic part segmentation, and so it suffers when the semantic part segmentation is not accurate, as shown in the first row of Figure~\ref{fig:failure_case}.
Second, our model struggles to predict camera poses that are rare in the training dataset. For instance, the bird in the second row in Figure~\ref{fig:failure_case} presents a rare case where the camera is located very close to the bird, which is not correctly predicted by our model.
Third, our model, including the semantic template, is trained in a fully data-driven way. It captures the major shape characteristics of each instance but ignores some details, e.g., the two wings of flying birds, and the legs of zebras or horses are not separated, as shown in Figure~\ref{fig:bird} and Figure~\ref{fig:zebra}.
We leave all these failure cases and limitations to future works.


\clearpage
\bibliographystyle{splncs04}
\bibliography{egbib}
\end{document}